\pdfoutput=1

\documentclass[11pt]{article}

\usepackage[ruled,vlined]{algorithm2e}
\usepackage{amsmath,amssymb,amsfonts}
\usepackage[noend]{algpseudocode}
\usepackage{bm}	
\usepackage{bbm}
\usepackage{graphicx}
\usepackage[final]{acl}
\usepackage{multirow}
\usepackage{soul}
\usepackage{times}
\usepackage{latexsym}
\usepackage{floatrow}
\usepackage{booktabs}
\usepackage{rotating}
\usepackage{multicol}
\usepackage{multirow}
\usepackage{fontawesome}
\usepackage{xcolor}

\usepackage[T1]{fontenc}

\usepackage[utf8]{inputenc}

\usepackage{microtype}

\usepackage{inconsolata}
\usepackage{enumitem}
\usepackage{array}

\newcolumntype{H}{>{\setbox0=\hbox\bgroup}c<{\egroup}@{}}

\newcommand{\mymethod}{\sc{AdpMixup}}
\definecolor{darkspringgreen}{rgb}{0.09, 0.45, 0.27}
\title{{\sc Adapters Mixup}: Mixing Parameter-Efficient Adapters to Enhance the Adversarial Robustness of Fine-tuned Pre-trained Text Classifiers}

\author{Tuc Nguyen \\
  Department of Computer Science \\
  Indiana University \\
  \texttt{nguyentuc1003@gmail.com} \\\And
  Thai Le \\
  Department of Computer Science \\
  Indiana University \\
  \texttt{tle@iu.edu} \\}

\begin{document}
\maketitle
\begin{abstract} 
Existing works show that augmenting the training data of pre-trained language models (PLMs) for classification tasks fine-tuned via parameter-efficient fine-tuning methods (PEFT) using both clean and adversarial examples can enhance their robustness under adversarial attacks. 
However, this adversarial training paradigm often leads to performance degradation on clean inputs and requires frequent re-training on the entire data to account for \textit{new, unknown attacks}. 
To overcome these challenges while still harnessing the benefits of adversarial training and the efficiency of PEFT, this work proposes a novel approach, called {\mymethod}, that combines two paradigms: \textit{(1) fine-tuning through adapters} and \textit{(2) adversarial augmentation via mixup} to dynamically leverage existing knowledge from a set of pre-known attacks for robust inference.
Intuitively, {\mymethod} fine-tunes PLMs with multiple adapters with both clean and pre-known adversarial examples and intelligently mixes them up in different ratios during prediction.
Our experiments show {\mymethod} achieves the best trade-off between training efficiency and robustness under both \textit{pre-known and unknown} attacks, compared to existing baselines on five downstream tasks across six varied black-box attacks and 2 PLMs. 
All source code will be available. 
\end{abstract}

\section{Introduction}
PEFT exemplified by adapter methods, offers a promising solution to mitigate fine-tuning costs for PLMs. PEFT involves injecting a small set of parameters into specific locations within a PLM, activating only these parameters while freezing the remainder during training. This approach significantly reduces the number of trainable parameters to as little as $0.1\%$ of the original count, while maintaining competitive performance on downstream tasks \cite{wang-etal-2022-adamix}. Adversarial training includes textual adversarial examples during training to enhance adversarial robustness for neural network models, including PLMs~\cite{goodfellow_explaining_2014,miyato2018virtual, zhu2019freelb}.
However, fine-tuning PLMs via this paradigm to be robust against various types of adversarial perturbations is computationally expensive due to the necessity of independently re-training the models with different perturbations to accommodate different types of attack methods. In addition, adversarial training often decreases performance on clean examples~\cite{xu2021robust}. 

 \begin{figure}[tb]
    \centering
    \includegraphics[width=0.995\textwidth]{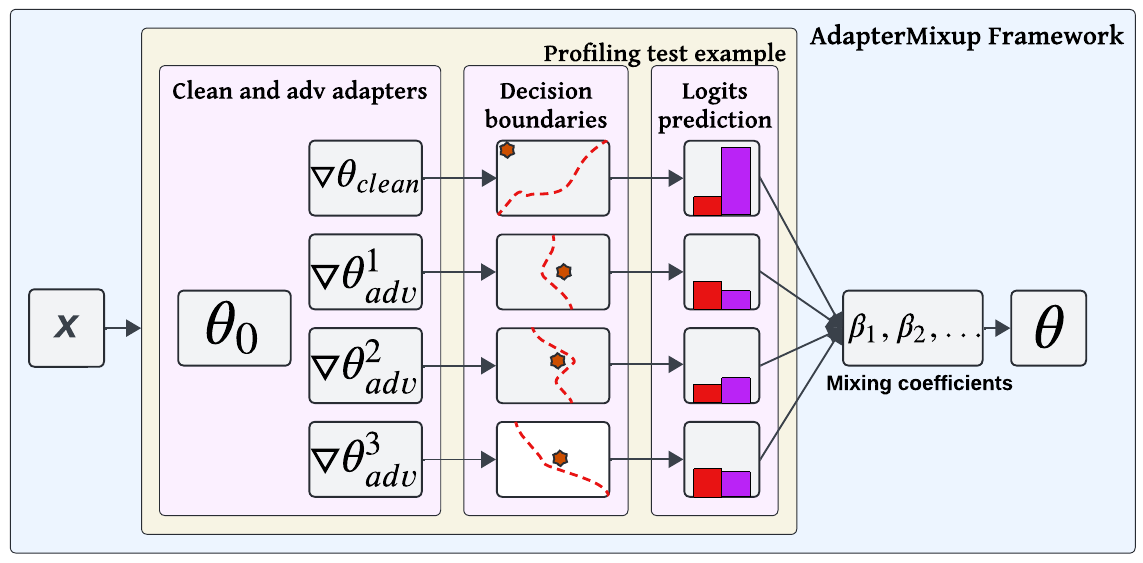}
    \caption{{\mymethod} Framework: Final model $\theta$ is achieved by dynamically mixing the adapter weights across clean and adversarial with different coefficients $\beta_1, \beta_2, \dots $. The \textcolor{red}{\textit{dash red lines}} are the decision boundaries of different fine-tuning models, that when mixed in a certain way can result in robust inference.
    }
    \label{adamixup_framework}
\end{figure}

To further improve adversarial training, \citet{miyato2018virtual} introduces Mixup, which trains a model on virtual examples constructed via linear interpolation between two random examples from the training set and their labels. Mixup also helps to improve model robustness under a variety of mixing styles, including mixing between original examples, between original examples and their adversarial examples, and between only adversarial examples~\cite{si2021better}.
However, Mixup shares the same inefficiency with adversarial training in practice as we need to retrain entire models every time we need to accommodate new types of attacks. It is also unknown how Mixup can be efficiently applied to fine-tuning PLMs via PEFT. In fact, 
there is limited number of research addressing PLMs's generalization capabilities and adversarial robustness~\cite{nguyen2024generalizability} when fine-tuned via PEFT, not to mention that many of existing defense methods are not specifically designed for PEFT, highlighting a critical gap in the literature. These observations prompt a crucial question: \textit{``How can we use PEFT with PLMs on downstream tasks that can achieve better trade-off among accuracy, adversarial robustness, and computational complexity and also withstand a variety of new, unknown attack methods?''}
To answer this question, we seek to investigate how to incorporate adversarial data augmentation training to improve PLMs' adversarial robustness without sacrificing performance on clean examples, while making minimal changes to complex PLMs during fine-tuning to minimize computational overhead with PEFT.

To tackle this, this work presents a novel approach, called {\mymethod}, that combines two key paradigms: \textit{(1) fine-tuning through PEFT, often referred to as adapters}~\cite{houlsby2019parameter, hu2021lora} and \textit{(2) adversarial augmentation via mixup}~\cite{miyato2018virtual}. Intuitively, {\mymethod} fine-tunes PLMs with multiple adapters with both clean and pre-known adversarial examples and mix them up in different ratios for robust inference (Fig.~\ref{adamixup_framework}). This new adapters mixup paradigm allows {\mymethod} to work well in practice when the attack methods by which possible adversarial examples are generated are unknown.


Our contributions are summarized as follows.
\vspace{-3pt}
\begin{enumerate}[leftmargin=\dimexpr\parindent-0.2\labelwidth\relax,noitemsep]
    \item Provide an analysis of the connections between data augmentation methods adversarial training, Mixup, and model augmentation methods including ModelSoup and PEFT via Adapters;
    \item Propose {\mymethod} that combines adversarial training, Mixup, and Adapters to achieve the best trade-off between training efficiency, and predictive performance under both clean and adversarial examples generated via pre-known and unknown attacks on five classification datasets;
    \item {\mymethod} also achieves the best trade-off in generalizability under both clean and adversarial examples, and superior efficient space and runtime complexity in practice.
    \item {\mymethod} also enables the profiling of potential adversarial examples by characterizing them into pre-known attacks, allowing more interpretable analysis of risk analysis in practice.
\end{enumerate}

\section{Related Work}
\subsection{Training with Data Augmentation}
Let denote $f(\cdot; \theta)$ a PLM parameterized by $\theta$, $(x_i, y_i)$ 
an arbitrary clean input.

\paragraph{Adversarial Training }~\citet{goodfellow_explaining_2014}. Adversarial augmentation training optimizes $\theta$ on both clean and adversarial examples to improve $f$'s adversarial robustness by minimizing the loss:
\setlength{\abovedisplayskip}{2pt}
\begin{equation}
 \alpha L(f(x; \theta), y){+}(1{-}\alpha) \max_{\delta \in S} L(f(x{+}\delta; \theta), y),
\label{eq:co_training_loss}
\end{equation}
\noindent where $\delta$ is the adversarial perturbation, $\alpha$ controls how much the loss $L$ is updated towards the adversarial and clean version of the training input $x$ and label $y$, and $S$ is the set of allowed perturbations.

\paragraph{Mixup.} 
Adversarial augmentation training helps enhance the adversarial robustness of NNs models. However, studies such as \citet{xie_adversarial_2019} observe a consistent instability, often leading to a reduction in the trained models' performance on clean examples. This is a result of the substantial gap between clean and adversarial examples that can introduce non-smooth learning trajectories during training~\cite{si2021better}. To address this, \textit{Mixup}~\cite{mixup} was proposed as a data augmentation method via linear interpolation to tackle a model's sensitivity to adversarial examples, and its instability in adversarial training. 
Mixup trains a model on virtual examples constructed via linear interpolation between two random examples from the training set and their labels~\cite{zhang2017mixup}.
While Mixup has been proposed as suitable for continuous data, its application to text data raises questions about its natural fit. Defining a process of "convex combination" between two texts is mathematically feasible, yet the resulting text may \textit{lack grammatical correctness or semantic coherence}. This challenges Mixup's utility as a viable method for enhancing the robustness of PLMs. 
In practice, \textit{Mixup also shares the same inefficiency with adversarial training as we need to fine-tune a PLM on entire datasets to accommodate new types of attacks}.

\subsection{Training with Model Augmentation}

\label{model_augmentation}
\paragraph{Model Soup.} Model Soup averages model weights of a pool of $k$ models $\{f_1, f_2, \dots, f_k \}$ to achieve better robustness without incurring runtime required to make $k$ inference passes as seen in classical ensemble learning~\cite{wortsman2022model}. In principle, Model Soup is similar to Stochastic Weight Averaging~\citep{izmailov_averaging_2018} which averages model weights along an optimization trajectory.
Particularly, given two model weights $\theta_1$ and $\theta_2$, Model Soup with a coefficient $\alpha{\in}[0, 1]$ results in a single model with parameters:
\setlength{\abovedisplayskip}{2pt}
\setlength{\belowdisplayskip}{2pt}
\begin{equation}
    \theta_\alpha = \alpha \theta_1 + (1-\alpha) \theta_2
    \label{model_soup}
\end{equation}
\paragraph{Adapter.}
Adapters or PEFT help fine-tuning PLMs on downstream tasks or with new domains efficiently~\cite{houlsby2019parameter, hu2021lora}. Some works such as \cite{pfeiffer2020adapterfusion} also propose to use not only one but also multiple adapters to further enhance the generalizability of the fine-tuned models on not one but multiple domains.
Given $f$ with a PLM parameter $\theta_0$, fine-tuning $f$ via adapters on two domains $D_1$ and $D_2$ results in two sufficiently small adapter weights $\triangledown \theta_1$ and $\triangledown \theta_2$, respectively. This corresponds to two distinct models $f(\cdot;\theta_0 + \triangledown \theta_1)$ and $f(\cdot; \theta_0 + \triangledown \theta_2)$ during inference. Since $\triangledown \theta_1$ and $\triangledown \theta_2$ are designed to be very small in size compared to $\theta_0$, this approach \textit{helps achieve competitive performance compared to fully fine-tuning all model parameters $\theta_0$ with only a small fraction of the cost}.

\section{Motivation}
\label{motivation}
In this section, we demonstrate how data augmenting methods are linked to the model augmentation methods, which underpins the rationale for our {\mymethod} framework.

\subsection{Adversarial Training versus Mixup}
\label{data_augmentation_training}
\paragraph{Adversarial Training} helps enhance the adversarial robustness of NNs by jointly optimizing $\theta$ on both clean and adversarial data following the Eq. (\ref{eq:co_training_loss}). 
When $\alpha{\leftarrow}1$, Eq. (\ref{eq:co_training_loss}) converges to conventional training on only clean examples, resulting in $\theta{\leftarrow} \theta_{\operatorname{clean}}$. When $\alpha{\leftarrow}0$, it converges to adversarial training with only adversarial examples, resulting in $\theta{\leftarrow} \theta_{\operatorname{adv}}$~\cite{madry_towards_2017}.

\paragraph{Mixup.} 
Mixup \cite{mixup} is used to regularize NNs $f$ to favor simple linear behavior in between training examples by training $f$ on convex combinations of pairs of examples and their labels. Exploiting this property of Mixup, \cite{si2021better} proposes to adapt Mixup to augment training examples by interpolating not only between clean but between clean and adversarial samples.
Given two pairs of samples $(x_i, y_i)$ and its adversarial sample 
$(x^*_i, y^*_i)$, their Mixup interpolation results in:
\begin{equation}
    \label{mixup_clean_adversarial}
    \begin{aligned}
        (\overline{x}, \overline{y}) &= \operatorname{Mixup} ((x_i; y_i), (x^*_i; y^*_i)) \\
        &= [\lambda x_i + (1-\lambda) x^*_i; \lambda y_i + (1- \lambda) y^*_i ],
    \end{aligned}
\end{equation}
where $\lambda$ is the interpolation coefficient. 
When $\lambda{=}1$, the Mixup produces only clean samples, resulting in a trained model with parameter $\theta{=}\theta_{clean}$. When $\lambda{=}0$, the Mixup produces only adversarial examples, resulting in the trained model with parameter $\theta{=}\theta_{adv}$. 
From Eq. (\ref{eq:co_training_loss}), \ul{Mixup with adversarial examples under $\lambda{\leftarrow}0$ or $\lambda{\leftarrow}1$ converges to adversarial training with $\alpha{\leftarrow}0$ and $\alpha{\leftarrow}1$, respectively}.

\subsection{ModelSoup on Adapter}
\label{modelsoup}
Model Soup is used to averaging weights of multiple fine-tuned models
improves generalization without increasing inference time~\cite{wortsman2022model}. However, in Model Soup, when the model weights are substantially different, averaging them would result in conflicting or contradicting information acquired during \textit{pre-training}, leading to poor performance.
Model Soup's authors also advocate the selection of sub-models in decreasing order of their validation accuracy on the same task for optimal results, showing that the sub-models should sufficiently converge or, they are close in parameter space, especially for PLMs~\cite{neyshabur2020being}.

To pursue a harmonized optimization trajectory, we want $\theta_1$ and $\theta_2$ to exhibit substantial similarity, differing only in a few parameters responsible for their expertise.
This is the case of Adapters, as we can decompose $\theta_1{=}\theta_0{+}\triangledown \theta_1$, $\theta_2{=}\theta_0{+}\triangledown \theta_2$ where the sizes of $\triangledown \theta_1$, $\triangledown \theta_2$ are minimal compared to $\theta_1$ or $\theta_2$ ($\S$\ref{model_augmentation}). 
Hence, \ul{this motivates us to adopt adapters to maximize the similarity in optimization trajectories between two sub-models, enabling the training of a merged model that is more competitive.}
Moreover, merging adapters are also more efficient, only requiring fine-tuning a small set of additional parameters $\triangledown \theta_1$, $\triangledown \theta_2$ and not the whole $\theta_1$ and $\theta_2$.
Therefore, to get a single language model that generalizes well on the two tasks following parametrized model:
\begin{equation}
 f(\cdot; \theta_0 + [\beta \triangledown \theta_1 + (1-\beta)\triangledown \theta_2])
\label{eq:adapter_mixup}
\end{equation}
where $\beta$ is the weighting factor when averaging the adapters. If $\beta=1$, the \emph{Model Soup on Adapters} boils down to the $\theta_1$ mode and conversely $\beta=0$ corresponds to the $\theta_2$ mode.

\section{{\mymethod}: Mixup of Adapters with Adversarial Training}
\label{adt_mixup_method}
From $\S$~\ref{motivation}, we learned that \ul{the mixed model achieved by Model Soup on the Adapters can achieve performance close to the model training with Mixup data augmentation}.
Therefore, instead of doing Mixup on text samples which is not intuitive in practice, {\mymethod} allows us to do Mixup on the model weight but still preserve the effectiveness of the data augmentation method.

Let's define two adapters $\theta_{clean}, \theta_{adv}$ trained on clean and adversarial data, respectively. We have model in \emph{clean mode}  $f(\cdot; \theta_0{+}\triangledown \theta_{clean})$, and $f(\cdot; \theta_0{+}\triangledown \theta_\textrm{adv})$ is the model in \emph{adversarial mode}.
Prediction after mixing the two adapters via Mixup can then be formulated as:
\setlength{\abovedisplayskip}{2pt}
\begin{equation}
 f(\cdot; \theta_0 + [\beta \triangledown \theta_{clean} + (1-\beta)\triangledown \theta_\textrm{adv}]),
\label{eq:model_soup}
\end{equation}
where $\beta$ is the Mixup coefficient. When $\beta{=}1$, {\mymethod} boils down to the \emph{clean mode} and conversely $\beta{=}0$ corresponds to the \emph{adversarial mode}.

\subsection{Choosing $\beta$ dynamically}
Given a PLM $\theta_0$, clean adapter $\triangledown \theta_{clean}$ and adversarial adapter $\triangledown \theta_{adv}$, we want to find the optimal $\beta$ for every sample during inference based on entropy to measure uncertainty.

Specifically, given $P_{clean}(x)$ is the probability of prediction of clean model $\theta_{clean}$ on example $x$. Then the entropy $H$ measures the expected information content of the prediction $P_{clean}(x)$ is computed following:
\begin{equation}
 H(P_{clean}(x)) = - \sum_{p(x)} p(x) \operatorname{log} (p(x)),
\label{entropy_equation}
\end{equation}
where $p(x)$ is probability prediction of $P_{clean}(x)$ over classification label.
Since the prediction $P_{clean}(x)$ will be close to the uniform distribution on adversarial example. Therefore, \ul{the entropy of the prediction of the clean model should be high on adversarial examples}. 
As a consequence, if the test samples are close to the clean set, $\beta$ should be close to 1, and vice versa if the test samples are close to the adversarial set, $\beta$ should be close to 0 (Eq.~\ref{eq:model_soup}).

\begin{figure}[t]
    \centering
    \includegraphics[width=0.85\textwidth]{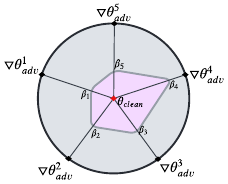}
    \caption{By choosing the coefficients $\beta$ dynamically, {\mymethod} allows us to profile the regions of combination weight. $\theta_0$ represents the pre-trained weight of the language model, while the gray area illustrates all possible combinations between clean and adversarial adapters. The pink area denotes the potential robust combinations of adapter weights.}
    \label{choose_beta}
\end{figure}

\subsection{Pre-knowing one adversarial attack} 
\paragraph{Measure how much clean adapter contributed to mix model.}
Let denote $H(P_{clean}(x_1))$, $H(P_{clean}(x_2))$, \dots, $H(P_{clean}(x_k)) $ is the set of entropy of clean prediction over 100 samples train clean dataset. $\operatorname{max_{clean}}$ and $\operatorname{min_{clean}}$ are the maximum and minimum entropy, respectively. With test example $x_i (i \in [0, k])$, we estimate the contribution of a clean adapter by the maximum normalization following:
\setlength{\abovedisplayskip}{2pt}
\setlength{\belowdisplayskip}{2pt}
\begin{equation}
 \alpha^{clean}_i = \frac{max_{clean} - H(P_{clean}(x_i))}{max_{clean} - min_{clean}}.
\label{maximum_normalization}
\end{equation}
and the mixed model is used to predict $x_i$ is computed following:
\begin{equation}
 \theta_i = \theta_0 + [\alpha^{clean}_i \triangledown\theta_{clean} + (1- \alpha^{clean}_i) \triangledown\theta_{adv}]
\label{mix_clean_model}
\end{equation}
Intuitively, if example $x_i$ is a clean example, $H(P_{clean}(x_i))$ will be low, $\alpha^{clean}_i$ will be high, and the mixed model $\theta_i$ will close with the clean model.
On the opposite, if example $x_i$ is an adversarial example, $H(P_{clean}(x_i))$ will be high, $\alpha^{clean}_i$ will be low, and the mixed model $\theta_i$ will close with the adversarial model. To summarize, \ul{$\alpha^{clean}_i$ controls how much a clean adapter contributes to the mixed model}.

\paragraph{Measure how much adversarial adapter contributed to mix model.}
Similarly, we compute $\alpha^{adv}_i$ by using minimum normalization with the set of entropy of the adversarial model over 100 adversarial training samples $H(P_{adv}(x_1))$, $H(P_{adv}(x_2))$, \dots, $H(P_{adv}(x_k))$. \ul{$\alpha^{adv}_i$ controls how much an adversarial adapter contributes to the mixed model}.
Then the mixed model is used to predict $x_i$ is computed following:
\begin{equation}
 \theta_i = \theta_0 + [\alpha^{adv}_i \triangledown\theta_{clean} + (1- \alpha^{adv}_i) \triangledown\theta_{adv}]
\label{mix_adv_model}
\end{equation}

\setlength{\abovedisplayskip}{2pt}
\setlength{\belowdisplayskip}{2pt}

In summary, let denote the mixing coefficient $\beta^i = (\alpha^{clean}_i + \alpha^{adv}_i)/2$, and average the RHS terms in the Eq. \ref{mix_clean_model} and \ref{mix_adv_model}:
\begin{equation}
 \theta_i = \theta_0 + [\beta^i \triangledown\theta_{clean} + (1- \beta^i) \triangledown\theta_{adv}]
\label{mixed_model}
\end{equation}

\subsection{Pre-knowing \textit{m} adversarial attack (m>1)} 
In scenarios where $m$ adversarial attacks are identified, we can construct $m$ pairs of clean and adversarial adapters for each example during inference.
If $m=2$, for every example on the evaluation set, we have two pairs of ($\triangledown\theta_{clean}$, $\triangledown\theta^1_{adv}$) and ($\triangledown\theta_{clean}$, $\triangledown\theta^2_{adv}$) with $2$ coefficient $(\beta_1, \beta_2)$. Specifically, for every sample $x_i(i\in [0,k])$, we have two mixed model which are formulated as:
\begin{align}
    \theta^1_i = \theta_0 + [\beta^i_1 \triangledown\theta_{clean} + (1- \beta^i_1) \triangledown\theta^1_{adv}] \\
    \theta^2_i = \theta_0 + [\beta^i_2 \triangledown\theta_{clean} + (1- \beta^i_2) \triangledown\theta^2_{adv}]
\end{align}
The final mixed model for sample $x_i$, utilizing one clean adapter and two known adversarial adapters, is computed as follows:
\begin{equation}
\begin{aligned}
    \theta_i &= (\theta^1_i + \theta^2_i) /2 =\theta_0 + \frac{(\beta^i_1 + \beta^i_2)}{2} \triangledown\theta_{clean} \\
    & + \frac{(1- \beta^i_1)}{2}\triangledown\theta^1_{adv} 
    + \frac{(1- \beta^i_2)}{2}\triangledown\theta^2_{adv}
\end{aligned}
\end{equation}

Generalizing for $m{>}2$, for every sample $x_i$, the final mixed model is computed as $\frac{\sum_{l=1}^{l=m} \theta^l_i}{m}$, where $\theta^l_i$ represents the prediction from the $l$-th adversarial adapter for sample $x_i$.
\begin{align}
    \theta_i &=\theta_0 + \frac{\sum_{l=1}^{l=m} \beta^i_l}{m} \triangledown\theta_{clean} +  \frac{\sum_{l=1}^{l=m}(1- \beta^i_l)}{m}\triangledown\theta^l_{adv} 
    \label{final_eq}
\end{align}
As a results, {\mymethod} utilizes the entropy of model predictions as a metric to quantify the contribution of each adapter, potentially impacting the final mixed model for every new incoming sample. The visualization of the profiling weight for each incoming input is depicted in Fig.~\ref{choose_beta}.

\section{Experiment Set-up} 
\label{sec:experiments}

\paragraph{Datasets and Models.} We evaluate the effectiveness of {\mymethod} on the GLUE benchmark dataset~\cite{wang2018glue} across 5 tasks. We present the average clean and adversarial accuracy on the test set. We evaluate our algorithm on the BERT~\cite{bert}, RoBERTa~\cite{RoBERTa} because they share the same architecture with the latter models and at the same time standard for benchmarking in existing works~\cite{si2021better}. We use the popular Houlsby adapter~\cite{houlsby2019parameter} as the PEFT method for efficient fine-tuning. We refer the readers to $\S$~\ref{data_length_statistic} and $\S$~\ref{sec:transfer_learning} (Appendix) for details of the benchmark datasets and the hyper-parameter configurations for fine-tuning our models, respectively.


\paragraph{Victim Models and Attack Methods.} Our experimentation involves two Victim Models, namely BERT~\cite{bert} and RoBERTa~\cite{RoBERTa}. 
We employ 4 different types of word-based text attackers, namely TextFooler (TF)~\cite{TextFooler}, PW~\cite{PWWS}, BAE~\cite{garg2020bae}, PS~\cite{zang2019word} and 2 types of character-based text attackers, namely DeepWordBug (DW)~\cite{gao2018black} and TextBugger (TB)~\cite{li2018textbugger}. They all observed superior effectiveness in attacking state-of-the-art PLMs while preserving as much as possible the original semantic meanings.
Notably, all the attack algorithms are black-box attackers--i.e., they can query the target models' predictions but their parameters or gradients, making our evaluation practical.
We refer the readers to \ref{textattack_configuration} for details of setting up the attackers.

\paragraph{Baselines.} We compare {\mymethod} with several baselines as follows.

\begin{itemize}[leftmargin=\dimexpr\parindent-0.2\labelwidth\relax,noitemsep,topsep=0pt]
    \item \textbf{\textit{Base Models}}:  $\theta_{\operatorname{clean}}$ (denoted as \textbf{\textit{CleanOnly}}), $\theta_{\operatorname{adv}}$ (denoted as \textbf{\textit{AdvOnly}}) are two models trained with only clean examples and with only adversarial examples, respectively.
    \item \textbf{\textit{AdvTrain}} ~\cite{miyato2018virtual} where we train a single model on the augmentation of clean and adversarial data.
    \item \textbf{\textit{ModelSoup}} \cite{wortsman2022model} where we average the weights of whole models independently trained on clean and adversarial data.
    \item \textbf{\textit{AdapterSoup}}~\cite{chronopoulou2023adaptersoup} where we average the weights of adapters independently trained on clean and adversarial data.
\end{itemize}

\section{Results}

\subsection{Defend Against Pre-Known Attacks}
Table \ref{avg_clean_adv_traditionadv_training} shows results when the attackers are known in advance. {\mymethod} outweighs augmentation methods in both data and model space in terms of averaged performance on clean and adversarial inputs across all types of attacks. Unlike adversarial training, {\mymethod}'s performance remains more or less the same with models trained with only clean examples. Although its performance under attacks was still below the model trained on \textit{only} adversarial examples, it achieves \textit{the best trade-off between with and without attacks} across all settings. Appendix \ref{detail_result_known_attack} provides more details.

\renewcommand{\tabcolsep}{3pt}
\begin{table}[t]
\footnotesize
\begin{tabular}{lcccc|ccc}
\toprule

& \multicolumn{1}{c}{\multirow{2}{*}{\textit{\textbf{Methods}}}} & \multicolumn{3}{c}{\textbf{RoBERTa}} & \multicolumn{3}{c}{\textbf{BERT}}   \\

\cmidrule(lr){3-5} \cmidrule(lr){6-8}
& & \multicolumn{1}{c}{Clean} & \multicolumn{1}{c}{Adv} & \multicolumn{1}{c}{Avg} & \multicolumn{1}{c}{Clean} & \multicolumn{1}{c}{Adv}  & \multicolumn{1}{c}{Avg}\\   

\cmidrule(lr){1-8}
\multirow{6}{*}{\begin{sideways} \textit{\textbf{Word-based}} \end{sideways}}
&\multicolumn{1}{l}{\textit{CleanOnly}} & 91.7 & 49.7 & 70.7 & 84.0 & 50.0 & 67.0\\ 

&\multicolumn{1}{l}{\textit{AdvOnly}} & 55.7 & 69.8 & 62.8 & 61.3 & 69.2 & 65.3\\ 

&\multicolumn{1}{l}{\textit{AdvTrain}} & 89.8 &  65.6& 77.7 & 81.8 & 61.1 & 71.5\\

&\multicolumn{1}{l}{\textit{ModelSoup}} & 77.1 & 59.8 & 68.5 & 70.0 & 54.4 & 62.2\\ 

&\multicolumn{1}{l}{\textit{AdapterSoup}}  & 90.6 &  66.7 & 78.7 & 82.4 & 64.6 & 73.5 \\ 

&\multicolumn{1}{l}{{\mymethod}}  & \textbf{91.6} & \textbf{71.4} & \textbf{81.5} & \textbf{83.2} & \textbf{66.4} & \textbf{74.8}\\













\cmidrule(lr){1-8}
\multirow{6}{*}{\begin{sideways} \textit{\textbf{Character-based}} \end{sideways}}
&\multicolumn{1}{l}{\textit{CleanOnly}} & 91.7 & 59.3 & 75.5 & 84.0 & 50.6 & 67.3\\ 

&\multicolumn{1}{l}{\textit{AdvOnly}}  & 53.1 & 78.7 & 65.9 & 50.0 & 73.2 & 61.6\\ 

&\multicolumn{1}{l}{\textit{AdvTrain}} & 89.2 & 71.6  & 80.4 & 82.3 & 66.9 & 74.6\\

&\multicolumn{1}{l}{\textit{ModelSoup}} & 69.2 & 64.8 & 67.0 & 67.4 & 57.4 & 62.4\\ 
&\multicolumn{1}{l}{\textit{AdapterSoup}}  & 90.4 & 72.9 & 81.7 & 82.3 &  70.2 & 76.3  \\

&\multicolumn{1}{l}{{\mymethod}}  & \textbf{91.8} & \textbf{76.6} & \textbf{84.2} & \textbf{83.1} & \textbf{71.0}  & \textbf{77.1}\\

\bottomrule
\end{tabular}
\caption{Average model performance over 5 datasets of independent clean and adversarial training, traditional adversarial training with RoBERTa, BERT under 6 different types of text adversarial attack. \textit{\textbf{Bold}: the best average under clean and adversarial examples.}}
\label{avg_clean_adv_traditionadv_training}
\end{table}

\renewcommand{\tabcolsep}{1pt}
\begin{table}[tb]
\caption{Cross-attack evaluation between character-based TextBugger (TB) and DeepWordBug (DW). \textit{\textbf{Bold}: the best average under clean and adversarial examples.}}
\label{cross_attack_character_base}
\footnotesize
\begin{tabular}{lcc|cc|cc|cc}
\toprule
\multicolumn{1}{c}{\multirow{3}{*}{\textit{\textbf{Methods}}}} & \multicolumn{4}{c}{\textbf{RoBERTa}} & \multicolumn{4}{c}{\textbf{BERT}}     \\
\cmidrule(lr){2-5}\cmidrule(lr){6-9}
& \multicolumn{2}{c}{TB$\rightarrow$DW} & \multicolumn{2}{c}{DW$\rightarrow$TB}  & \multicolumn{2}{c}{TB$\rightarrow$DW} & \multicolumn{2}{c}{DW$\rightarrow$TB} \\
\cmidrule(lr){2-3} \cmidrule(lr){4-5} \cmidrule(lr){6-7} \cmidrule(lr){8-9}

& \multicolumn{1}{c}{Clean} & \multicolumn{1}{c}{Adv}  & \multicolumn{1}{c}{Clean} & \multicolumn{1}{c}{Adv}  &
\multicolumn{1}{c}{Clean} & \multicolumn{1}{c}{Adv}  &
\multicolumn{1}{c}{Clean} & \multicolumn{1}{c}{Adv} \\

\cmidrule(lr){1-9}

\multicolumn{1}{l}{\textit{CleanOnly}} & 91.7 &58.5 & 91.7 & 60.2 & 84.0 &46.0 & 84.0 & 55.0\\ 
\multicolumn{1}{l}{\textit{AdvOnly}} &53.5  &67.2& 52.7 & 71.0 & 42.6 &72.3& 57.4 & 69.6 \\ 

\multicolumn{1}{l}{\textit{AdvTrain}}  & 89.0 &63.9  &89.4  &67.6 & 82.2 &66.3& 82.4 &66.9\\ 

\multicolumn{1}{l}{\textit{ModelSoup}} & 68.9 &{53.5} & 69.5 &{54.9}&  66.9 &{56.1} & 67.9 &{53.3}\\ 

\multicolumn{1}{l}{\textit{AdapterSoup}} & 89.2 & 67.6 & 90.5 & 70.7 & 81.2 & 70.4 & 82.5 & 68.9\\ 

\multicolumn{1}{l}{{\mymethod}} & \textbf{91.7} &\textbf{68.0} & \textbf{91.8} &\textbf{71.3}  & \textbf{82.5}&\textbf{71.1} & \textbf{82.8} &\textbf{69.4}\\ 

\bottomrule
\end{tabular}
\end{table} 

\renewcommand{\tabcolsep}{1.5pt}
\begin{table}[tb]
\footnotesize
\begin{tabular}{lcHcc|Hcc|Hcc|Hcc}
\toprule
&\multicolumn{1}{c}{\multirow{1}{*}{\textit{\textbf{Pre-Known Atk}}}} & \multicolumn{3}{c}{TF}  & \multicolumn{3}{c}{BA} & \multicolumn{3}{c}{PW}  & \multicolumn{3}{c}{PS} \\
\cmidrule(lr){3-5} \cmidrule(lr){6-8} \cmidrule(lr){9-11} \cmidrule(lr){12-14}

& \textbf{Target Atk} & \multicolumn{1}{H}{Clean}& \multicolumn{1}{c}{PS} & \multicolumn{1}{c}{PW} &\multicolumn{1}{H}{Clean}& \multicolumn{1}{c}{PS} & \multicolumn{1}{c}{PW} &\multicolumn{1}{H}{Clean}& \multicolumn{1}{c}{TF} & \multicolumn{1}{c}{BA} &\multicolumn{1}{H}{Clean}& \multicolumn{1}{c}{TF} & \multicolumn{1}{c}{BA}\\

\cmidrule(lr){1-14}
\multirow{5}{*}{\begin{sideways} \textit{\textbf{RoBERTa}} \end{sideways}}
&\multicolumn{1}{l}{\textit{CleanOnly}} & 91.7 &50.3  &56.1 & 91.7&50.3 &56.1 & 91.7&46.7 &45.4 & 91.7&46.7 &45.4 \\ 

&\multicolumn{1}{l}{\textit{AdvOnly}} & 54.5 &65.5 &68.2 & 53.0 &64.4 &66.4 & 55.6 &66.8 & 69.0 & 59.6 &62.8 &63.5\\ 

&\multicolumn{1}{l}{\textit{AdvTrain}}  & 90.0 &60.2  &62.8 & 89.5 &59.7 &61.2 & 90.0&63.3 & 62.7 & 90.1&57.7 &55.9 \\ 

&\multicolumn{1}{l}{\textit{ModelSoup}} & 84.7 &{56.0}  &{56.2}& 82.6 &{54.7}  &{55.8} & 82.5 &{53.5} & {52.5}& 76.7 &{50.8} &{50.6} \\ 

&\multicolumn{1}{l}{\textit{AdapterSoup}} & 90.9 & 63.7  & 65.7 & 90.7 & 63.6 & 65.6 & 90.9 & 65.5  & 69.5 & 90.0 & 61.5 & 62.5 \\ 

&\multicolumn{1}{l}{{\mymethod}} & \textbf{91.6}&\textbf{65.1} &\textbf{67.2}  & \textbf{91.6}&\textbf{64.5} &\textbf{67.1} &  \textbf{91.5}&\textbf{66.4} &\textbf{70.4} & \textbf{91.8} & \textbf{62.0} &\textbf{63.5}\\ 

\cmidrule(lr){1-14}
\multirow{5}{*}{\begin{sideways} \textit{\textbf{BERT}} \end{sideways}}
&\multicolumn{1}{l}{\textit{CleanOnly}} & 84.0 &51.3 &51.1& 84.0& 51.3 &51.1 & 84.0 &53.5 &45.8 & 84.0 &53.5 &45.8\\ 

&\multicolumn{1}{l}{\textit{AdvOnly}} & 48.3 &63.6 &63.5 & 73.9 &57.7 &56.4& 51.1&67.6 & 65.8 & 71.7 & 58.1 &57.2 \\ 

&\multicolumn{1}{l}{\textit{AdvTrain}} & 81.4 & 54.8 &58.7& 82.7 & 52.5 & 51.9 &  80.5&62.1& 53.3& 82.4 & 56.5& 51.4 \\ 

&\multicolumn{1}{l}{\textit{ModelSoup}} & 70.3 &{51.1} &{51.4}& 66.7&{45.6} & {43.9}& 70.1 & {53.0}& {50.2}& 74.0 &{47.4} &{45.6}\\ 

&\multicolumn{1}{l}{\textit{AdapterSoup}} &  82.6 & 59.3 & 62.0 & 83.2 & 55.4 & 54.3 & 81.5 & 64.6 & 62.9 & 82.2 & 57.1  & 55.9\\ 

&\multicolumn{1}{l}{{\mymethod}} & \textbf{83.2}&\textbf{61.3} & \textbf{63.6}& \textbf{83.5}  & \textbf{57.1} & \textbf{55.5} & \textbf{82.6} &\textbf{65.1} & \textbf{63.9}& \textbf{83.3} & \textbf{58.1}& \textbf{57.6}\\ 

\bottomrule
\end{tabular}
\caption{Cross attack evaluation among word-based methods. \textit{\textbf{Bold}: the best average under clean and adversarial examples.}}
\label{cross_attack_word_base}
\end{table}


\noindent \textbf{\textit{Analysis \#1: Mixing two \ul{whole, large} independent clean and adversarial models results in a significant decline in both generalization and adversarial robustness.}} This could happen because such independent models trained on datasets of different distributions may not converge to the same optimal trajectory. Thus, when combining them, the final mixed \textit{ModelSoup} model will not represent the optimal solution, as also shown in \cite{wortsman2022model}. This confirms our analysis in $\S$\ref{modelsoup}.

\noindent \textbf{\textit{Analysis \#2: {\mymethod} achieves better results on RoBERTa compared with BERT.}} {\mymethod} exhibits a larger decline in both clean and adversarial robustness on BERT compared to RoBERTa. This discrepancy may be attributed to the size of the adapter, which is 64 for RoBERTa, considerably smaller than the 256 in BERT. Consequently, RoBERTa allocates a smaller portion of weights across clean and adversarial classifiers compared to BERT. This limited weight sharing enables RoBERTa to achieve competitive performance compared to ensemble learning, as discussed in $\S$~\ref{modelsoup}.

\subsection{Defend with m=1 Pre-Known Attack}
In this scenario, we train the adapter on one adversarial dataset which is generated by one type of adversarial attack, and then evaluate its performance on adversarial datasets which are generated by different adversarial attacking algorithms. 

\vspace{2pt}
\noindent \textbf{Intra-type Settings.} Table~\ref{cross_attack_character_base} and \ref{cross_attack_word_base} present the average clean and adversarial robustness scores between character-based, and word-based across 5 downstream tasks. Due to computational limitations, in Table~\ref{cross_attack_word_base}, for each word-based attack method, we randomly selected two word-based methods as the target attacks. 
Overall, {\mymethod} achieves the best trade-off performance with and without attacks in utilizing the pre-known adversarial knowledge of one attacker to defend another unknown one.
We refer readers to Appendix~\ref{detailed_cross_attack_m_1} for detailed results.

\vspace{2pt}
\noindent \textbf{Inter-type Settings.} Table~\ref{group_avg_cross_attack} shows the model performance when trained on character-based adversarial datasets and evaluated on word-based adversarial datasets, and vice versa. Overall, injecting knowledge of adapters learned from character adversarial perturbations makes better improvement performance on word-based adversarial examples compared to knowledge learned from word-based.

\renewcommand{\tabcolsep}{1.2pt}
\begin{table}[tb]
\footnotesize
\begin{tabular}{lcccccc|ccc}
\toprule

&\multicolumn{1}{c}{\multirow{3}{*}{\textit{\textbf{Attacker}}}} &  \multicolumn{5}{c}{\textbf{Charac$\rightarrow$Word}} &  \multicolumn{3}{c}{\textbf{Word$\rightarrow$Charac}} \\
\cmidrule(lr){3-7} \cmidrule(lr){8-10}

&  & \multicolumn{1}{c}{Clean} & \multicolumn{1}{c}{TF} & \multicolumn{1}{c}{BAE} & \multicolumn{1}{c}{PS} & \multicolumn{1}{c}{PW}& \multicolumn{1}{c}{Clean}& \multicolumn{1}{c}{DW} & \multicolumn{1}{c}{TB} \\

\cmidrule(lr){1-10}
\multirow{5}{*}{\begin{sideways} \textit{\textbf{RoBERTa}} \end{sideways}}
&\multicolumn{1}{l}{\textit{CleanOnly}} & 91.7 &  46.8 & 45.6 & 50.3 & 56.1 & 91.7 &58.5 &60.0 \\ 

&\multicolumn{1}{l}{\textit{AdvOnly}} & 53.1 & 62.0 & 58.2& 60.8 & 64.7  & 55.7 & 64.9 & 66.4  \\ 

&\multicolumn{1}{l}{\textit{AdvTrain}}  & 89.2 & 56.2 & 51.8 & 56.1 & 65.4  & 89.8 & 62.4 & 62.8\\ 

&\multicolumn{1}{l}{\textit{ModelSoup}} & 69.2 & 52.8 & 43.5 & 46.1 & 47.4 & 81.6& 57.2& 58.0 \\ 

&\multicolumn{1}{l}{\textit{AdapterSoup}} & 91.4 & 59.8 & 60.1 & 59.6 & 64.4 & 90.6 & 63.1 & 64.2 \\ 

&\multicolumn{1}{l}{{\mymethod}} & \textbf{91.8} & \textbf{61.6} & \textbf{61.5}  & \textbf{60.7} & \textbf{64.4} &\textbf{91.6}& \textbf{64.5} & \textbf{66.1}\\ 

\cmidrule(lr){1-10}
\multirow{5}{*}{\begin{sideways} \textit{\textbf{BERT}} \end{sideways}}
&\multicolumn{1}{l}{\textit{CleanOnly}} &  84.0 & 53.5 & 45.8 &57.4 & 51.1 &84.0& 46.0 & 55.0 \\ 

&\multicolumn{1}{l}{\textit{AdvOnly}} & 50.0 & 63.2 & 65.3 & 65.7 & 65.5 &61.3& 63.9 &63.7\\ 

&\multicolumn{1}{l}{\textit{AdvTrain}} & 82.3 &  56.8 &  52.7 & 56.3 & 59.9 &81.8& 57.8 & 60.8  \\ 

&\multicolumn{1}{l}{\textit{ModelSoup}} & 67.4 & 52.0 & 49.0 & 47.2 & 49.6 &70.0& 47.9 & 51.0\\ 

&\multicolumn{1}{l}{\textit{AdapterSoup}} & 82.3 & 61.9 & 61.5 & 63.6 &  62.1 & 82.4 & 59.8 &  61.0\\ 

&\multicolumn{1}{l}{{\mymethod}} & \textbf{83.1} & \textbf{62.8} & \textbf{62.0} & \textbf{64.2} & \textbf{65.0} &\textbf{83.2}& \textbf{63.0} & \textbf{63.1}\\ 

\bottomrule
\end{tabular}
\caption{Average Cross Pre-Known Character and Pre-Known Word attack. \textit{In \textbf{bold} means best average performance with and without attack.}}
\label{group_avg_cross_attack}
\end{table}

\subsection{Defend with m{>}1 Pre-Known Attacks}
Table \ref{know_m_cross_attack_word_base} shows the cross-attack evaluations when increasing the number of known adversarial attacks.

\renewcommand{\tabcolsep}{2.0pt}
\begin{table}[tb]
\footnotesize
\begin{tabular}{lccc|cc|cc}
\toprule
&\multicolumn{1}{c}{\multirow{2}{*}{\textit{\textbf{Attacker}}}} & \multicolumn{2}{c}{\textbf{m=1}} & \multicolumn{2}{c}{\textbf{m=2}} & \multicolumn{2}{c}{\textbf{m=3}} \\
\cmidrule(lr){3-4} \cmidrule(lr){5-6} \cmidrule(lr){7-8}
& & \multicolumn{1}{c}{Clean} & \multicolumn{1}{c}{Adv} & \multicolumn{1}{c}{Clean} & \multicolumn{1}{c}{Adv} & \multicolumn{1}{c}{Clean} & \multicolumn{1}{c}{Adv}\\

\cmidrule(lr){1-8}
\multirow{5}{*}{\begin{sideways} \textit{\textbf{RoBERTa}} \end{sideways}}
&\multicolumn{1}{l}{\textit{CleanOnly}} &91.7 &49.7 &  91.7 &49.7 & 91.7 & 49.7 \\ 

&\multicolumn{1}{l}{\textit{AdvOnly}} & 55.7& 65.8 & 55.8 & 66.8\textcolor{darkspringgreen}{\faArrowUp} & 55.1 &68.6\textcolor{darkspringgreen}{\faArrowUp} \\ 

&\multicolumn{1}{l}{\textit{AdvTrain}}  & 89.9 & 60.4 & 90.2 & 61.4\textcolor{darkspringgreen}{\faArrowUp} & 86.1 & 64.0\textcolor{darkspringgreen}{\faArrowUp}\\ 

&\multicolumn{1}{l}{\textit{ModelSoup}} & 81.6 & 53.8 & 67.5 & 49.7\textcolor{red}{\faArrowDown} & 60.5 & 46.4\textcolor{red}{\faArrowDown} \\ 

&\multicolumn{1}{l}{\textit{AdapterSoup}} & 90.6 & 64.7 & 87.3 & 62.1\textcolor{red}{\faArrowDown} & 76.4 & 59.8\textcolor{red}{\faArrowDown} \\ 

&\multicolumn{1}{l}{{\mymethod}} & \textbf{91.6} & \textbf{65.7}  & \textbf{89.6} & \textbf{68.0}\textcolor{darkspringgreen}{\faArrowUp} & \textbf{91.6} & \textbf{69.1}\textcolor{darkspringgreen}{\faArrowUp}\\ 

\cmidrule(lr){1-8}
\multirow{5}{*}{\begin{sideways} \textit{\textbf{BERT}} \end{sideways}}
&\multicolumn{1}{l}{\textit{CleanOnly}} & 84.0 & 49.9 & 84.0 & 49.9 & 84.0 &49.9 \\ 

&\multicolumn{1}{l}{\textit{AdvOnly}} & 61.1 & 61.4 & 41.5 & 64.0\textcolor{darkspringgreen}{\faArrowUp} & 44.3 & 65.4\textcolor{darkspringgreen}{\faArrowUp}\\ 

&\multicolumn{1}{l}{\textit{AdvTrain}}  & 81.4 & 55.0 & 80.0 & 57.2\textcolor{darkspringgreen}{\faArrowUp} & 77.8 & 59.3\textcolor{darkspringgreen}{\faArrowUp} \\ 

&\multicolumn{1}{l}{\textit{ModelSoup}} & 70.8 & 48.5 & 63.5 & 44.6\textcolor{red}{\faArrowDown} & 52.9 & 42.6\textcolor{red}{\faArrowDown} \\ 

&\multicolumn{1}{l}{\textit{AdapterSoup}} &  82.4 & 59.1 & 76.5 & 56.5\textcolor{red}{\faArrowDown} & 68.9 & 52.9\textcolor{red}{\faArrowDown}\\ 

&\multicolumn{1}{l}{{\mymethod}} & \textbf{83.3} & \textbf{60.3} & \textbf{80.3} & \textbf{60.8}\textcolor{darkspringgreen}{\faArrowUp} & \textbf{83.6} & \textbf{63.5}\textcolor{darkspringgreen}{\faArrowUp}\\ 

\bottomrule
\end{tabular}
\caption{Cross Attack Evaluation between Word-based methods when knowing more than one adversarial attack. \textit{In \textbf{bold} means best average performance with and without attack.} \textcolor{darkspringgreen}{\faArrowUp}/\textcolor{red}{\faArrowDown} denotes the increase/decrease from preceding $m$ pre-known attacks.}
\label{know_m_cross_attack_word_base}
\end{table}

\noindent \textbf{\textit{Analysis \#1: {\mymethod} effectively utilizes pre-known attacks to defend against unknown ones.}} Increasing the number of pre-known attacks $m$ from 1 to 3 leads to an improvement in robustness for RoBERTa under attacks from $65.8\%$ with $m=1$ to $66.8\%$ and $68.6\%$ with $m$ is equal to 2 and 3, respectively. 
In addition, {\mymethod} demonstrates sustained competitive performance on clean data while benefiting from enhanced performance on adversarial datasets when subjected to an increased number of text adversarial attacks. This resilience may be attributed to {\mymethod} strategy of selecting different weight configurations for the clean and adversarial adapters, effectively harnessing insights from adversarial adapters to boost overall model performance.

\noindent \textbf{\textit{Analysis \#2: Model generalization decreases when increasing the number of known $m$ attacks.}}
When $m$ increases from 1 to 3, \textit{AdvOnly}, \textit{AdvTrain}, \textit{ModelSoup}, \textit{AdapterSoup} show a significant drop in model generalization on both RoBERTa and BERT (Table~\ref{know_m_cross_attack_word_base}).
This may stem from adversarial training inducing a shift in data distribution. Adversarial examples often deviate from the statistical distribution of clean data. 
Consequently, the training process might prioritize learning features and patterns specific to adversarial examples, diverging from the underlying data distribution of clean samples~\cite{goodfellow2014explaining}.


\section{Discussion}

\paragraph{Flexibility.} Compared to the baselines, {\mymethod} is able to leverage recent state-of-the-art PEFT methods with superior performance compared to Adapters \cite{houlsby2019parameter}, such as LoRA \cite{hu2021lora}, AdaMix \cite{wang-etal-2022-adamix}. Consequently, {\mymethod} exhibits modular properties, enabling the defense against new types of adversarial attacks by conveniently training a new adapter corresponding to the specific new attack and subsequently merging them. 

\begin{figure}[tb]
    \centering
    \includegraphics[width=0.85\textwidth]{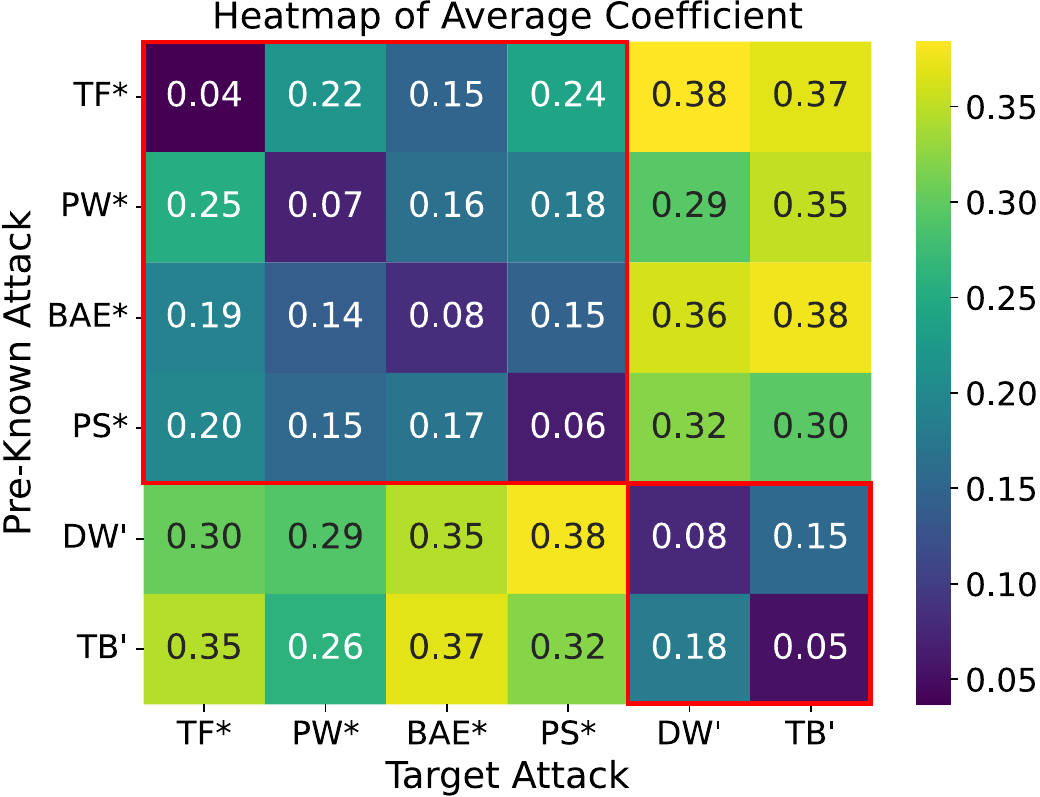}
    \caption{Average coefficient $\beta$ of \texttt{\mymethod} with $m=1$ pre-known attack during inference on 100 test examples with RoBERTa against different attack methods. The lower the score, the more the adversarial adapter weight contributes to the mixed models. * and ' denote word-based and character-based attacks, respectively. \textcolor{red}{Red rectangles} denote attacks of the same type (word or character-based).}
    \label{heat_map_average_coefficient}
\end{figure}

\paragraph{Profiling adversarial examples via analyzing $\beta$.}
Fig.~\ref{heat_map_average_coefficient} shows heatmap of average mixing coefficient $\beta$ with $m{=}1$ pre-known attack. For every pre-known attack, we use {\mymethod} to compute the set of mixing coefficient $\beta$ to be used during inference on 100 samples generated from 6 types of target attack. 
Overall, the weights from word-based adversarial adapters contribute \textit{more} to the final mixed model than the those from character-based adversarial adapters when the target attack is word-based. Similar observations can be made with character-based attacks. In other words, {\mymethod} enables interpretable and intuitive characterization of unknown target attacks by attributing them to the suitable set of pre-known attacks.

\begin{figure}[tb]
    \centering
    \includegraphics[width=0.9\textwidth]{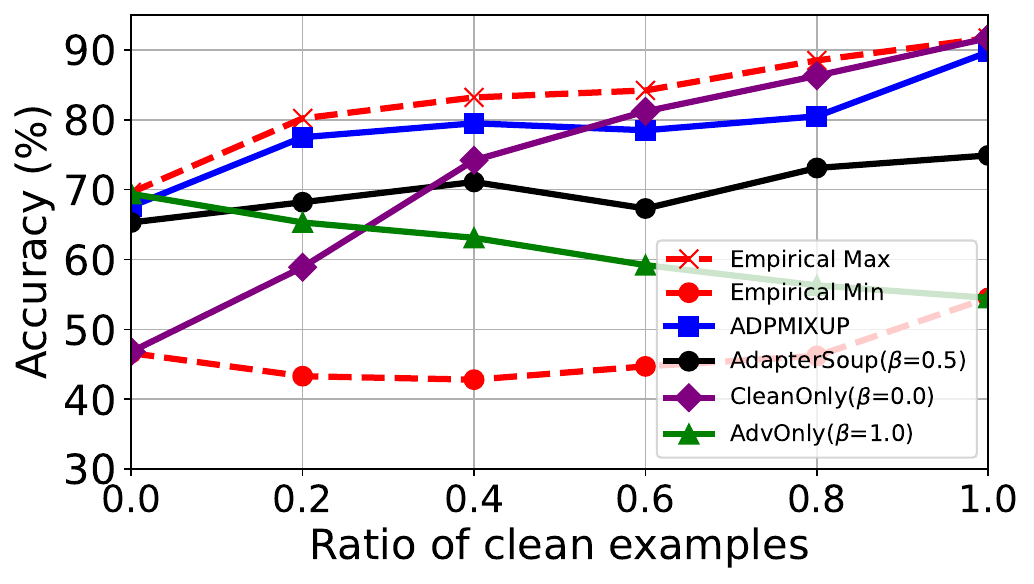}
    \caption{Average model accuracy (clean and adversarial) across 5 domain tasks under m{=}1 pre-known attack method at various ratios of clean examples.}
    \label{avg_adpmixup_max_min_textfooler_roberta}
\end{figure}

\paragraph{Empirical Min and Empirical Max.}
Fig.~\ref{avg_adpmixup_max_min_textfooler_roberta} shows the average accuracy of {\mymethod} across 5 tasks with m{=}1 pre-known attack under various ratios of clean examples. For each specific ratio of clean examples, we scan all the the coefficients $\beta{\in}[0,1]$ with step size of 0.1 to find ones that result in the best and the worst performance.
AdapterSoup's performance ($\beta$=0.5 fixed) remains more stable than CleanOnly ($\beta$=0.0 fixed) and AdvOnly ($\beta$=1.0 fixed) as the ratio of clean examples increases. However, their performance are very far away from the empirical optimal performance.
{\mymethod} (dynamic $\beta$) automatically finds the suitable coefficient $\beta$, \textit{achieving much closer performance to the empirical optimal $\beta$}. This further demonstrates the effectiveness of {\mymethod}'s intuition and design of using entropy to dynamically calculate the best set of $\beta$ for robust inference.

\begin{figure}[tb]
    \centering
    \includegraphics[width=0.9\textwidth]{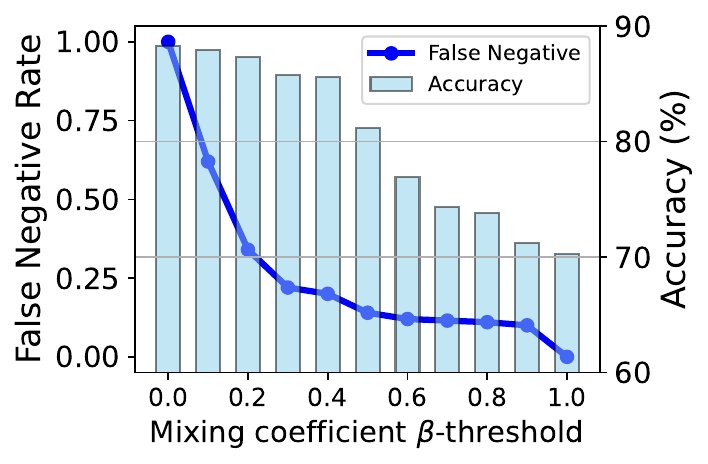}
    \vspace{-10pt}
    \caption{Trade-off between predictive accuracy (\textit{bar}) and false negative rate in detecting adversarial examples (\textit{line}) of {\mymethod} with RoBERTa, assuming a \textit{conservative} 15\% ratio of adversarials out of 1K test inputs.}
    \label{inference_time_acc_tradeoff}
\end{figure}

\paragraph{Theoretical and empirical computational complexity.}
Table \ref{detail_cost_of_training} shows that {\mymethod} has near optimal complexities in terms of space and training time compared to the baselines due to marginal additional computations required to accommodate $m$ adapters. During inference, {\mymethod} requires $\mathbf{\mathcal{O}(m)}$ complexity to calculate all the mixing coefficients $\beta$ (Eq. \ref{final_eq}). However, in practice, {\mymethod} does \textit{not always} need to use all $m$ adapter heads if the input is clean, as there are usually much less number of adversarial examples. Thus, to further reduce the runtime during inference, we can set a threshold on calculated $\beta$ on the clean adapter head to detect if an input is a potential adversarial example. With coefficient threshold $\beta$ is set to 0.4, {\mymethod} has a false negative rate on detecting adversarial examples of 0.25, and the accuracy of {\mymethod} only drops $2.7\%$ ($88.3\%{\rightarrow}85.6\%$) (Fig. \ref{inference_time_acc_tradeoff}). This help reduces the runtime significantly as it only needs to use all $m$ adapters \textit{maximum} about 30\% of the time. This makes {\mymethod}'s runtime complexity ($\sim\mathcal{O}(0.3m)$) closer or even better than $\mathcal{O}(1)$ when $m{\leq}3$, making {\mymethod}'s overall complexity practical in real-life, given that the ratio of adversarial examples can be much less than 15\% as used in Fig. \ref{inference_time_acc_tradeoff}.

\renewcommand{\tabcolsep}{1pt}
\begin{table}[tb]
\caption{Theoretical complexity during training and inference on a single example. $\theta_i$ represents an individual model. $m$ is the number of pre-known attacks. $\theta_0$ is the pre-trained weight that is shared across models. $\triangledown \theta_i$ is the adapter trained for task $i$-th.}
\begin{center}
    \footnotesize
    \begin{tabular}{lcccc} 
    \toprule
     \textbf{Method} & \textbf{Notation}  & \textbf{Training} & \textbf{Space} & \textbf{Inference} \\
     \midrule
     \textit{CleanOnly} & $ f(x, \theta)$ & $\mathcal{O}(1)$ & $\mathcal{O}(1)$ & $\mathcal{O}(1)$ \\
     \textit{AdvOnly} & $ f(x, \theta')$ & $\mathcal{O}(1)$ & $\mathcal{O}(1)$ & $\mathcal{O}(1)$ \\
     \textit{ModelSoup} & $ f(x,  \cup_{i=1}^m \theta_i)$ & $\mathcal{O}(m)$ &  $\mathcal{O}(m)$ & $\mathcal{O}(1)$  \\ 
     \textit{AdvTrain} & $f(x, \cup_{i=1}^m \theta'_i)$ & $\mathcal{O}(m)$ & $\mathcal{O}(1)$ & $\mathcal{O}(1)$  \\
     \textit{AdapterSoup} & $f(x, \theta_0 \cup_{i=1}^m \triangledown \theta_i )$ & ${\sim}\mathcal{O}(1)$ & ${\sim}\mathcal{O}(1)$ & ${\sim}\mathcal{O}(1)$  \\
     \cmidrule(lr){1-5}
     {\mymethod} & $f(x,  \theta_0{\cup_{i=1}^n}\triangledown \theta_i)$ & ${\sim}\mathbf{\mathcal{O}(1)}$ & ${\sim}\mathbf{\mathcal{O}(1)}$ & $\mathbf{\mathcal{O}(m)}$  \\
     \bottomrule
    \end{tabular}
\end{center}
\label{detail_cost_of_training}
\end{table}

\section{Conclusion}
This work provides a new framework for improving model generalization and robustness of PLMs under adversarial attacks by combining adversarial augmentation via Mixup and parameter-efficient fine-tuning via adapters.
Our findings highlight the utility of adapters in empowering PLMs to achieve competitive performance in terms of generalization and robustness under both pre-known and unknown adversarial attacks with minimal additional computational complexity.
Additionally, {\mymethod} provides extra interpretability into profiling and analyzing potential adversarial examples in practice.

\section*{Limitation}
Primarily, {\mymethod} use weight average \cite{wortsman2022model} to compute the weight of the final mixed model based on the mixing coefficient $\beta$.  Consequently, future works could investigate the applicability of our findings to these alternative model merging approaches.
Furthermore, our exploration focused solely on one BERT and RoBERTa on the natural language understanding tasks. 
As a result, a valuable avenue for future research would involve extending our analysis to encompass the emerging text generation tasks, particularly within the context of the current transformer-based language model like complex GPT-family models.

\section*{Broader Impacts and Ethics Statement}
We expect no ethical concerns regarding the artifacts and real-life applications of this work.

\bibliography{main}

\begin{thebibliography}{28}
\expandafter\ifx\csname natexlab\endcsname\relax\def\natexlab#1{#1}\fi

\bibitem[{Chronopoulou et~al.(2023)Chronopoulou, Peters, Fraser, and Dodge}]{chronopoulou2023adaptersoup}
Alexandra Chronopoulou, Matthew~E Peters, Alexander Fraser, and Jesse Dodge. 2023.
\newblock Adaptersoup: Weight averaging to improve generalization of pretrained language models.
\newblock In \emph{EACL}.

\bibitem[{Devlin et~al.(2019)Devlin, Chang, Lee, and Toutanova}]{bert}
Jacob Devlin, Ming-Wei Chang, Kenton Lee, and Kristina Toutanova. 2019.
\newblock \href {https://www.aclweb.org/anthology/N19-1423} {{BERT}: Pre-training of deep bidirectional transformers for language understanding}.
\newblock In \emph{NAACL}.

\bibitem[{Gao et~al.(2018)Gao, Lanchantin, Soffa, and Qi}]{gao2018black}
Ji~Gao, Jack Lanchantin, Mary~Lou Soffa, and Yanjun Qi. 2018.
\newblock Black-box generation of adversarial text sequences to evade deep learning classifiers.
\newblock In \emph{IEEE Security and Privacy Workshops}.

\bibitem[{Garg and Ramakrishnan(2020)}]{garg2020bae}
Siddhant Garg and Goutham Ramakrishnan. 2020.
\newblock Bae: Bert-based adversarial examples for text classification.
\newblock In \emph{EMNLP}.

\bibitem[{Goodfellow et~al.(2015{\natexlab{a}})Goodfellow, Shlens, and Szegedy}]{goodfellow_explaining_2014}
Ian~J Goodfellow, Jonathon Shlens, and Christian Szegedy. 2015{\natexlab{a}}.
\newblock Explaining and harnessing adversarial examples.
\newblock In \emph{ICLR}.

\bibitem[{Goodfellow et~al.(2015{\natexlab{b}})Goodfellow, Shlens, and Szegedy}]{goodfellow2014explaining}
Ian~J Goodfellow, Jonathon Shlens, and Christian Szegedy. 2015{\natexlab{b}}.
\newblock Explaining and harnessing adversarial examples.
\newblock \emph{ICLR}.

\bibitem[{Houlsby et~al.(2019)Houlsby, Giurgiu, Jastrzebski, Morrone, De~Laroussilhe, Gesmundo, Attariyan, and Gelly}]{houlsby2019parameter}
Neil Houlsby, Andrei Giurgiu, Stanislaw Jastrzebski, Bruna Morrone, Quentin De~Laroussilhe, Andrea Gesmundo, Mona Attariyan, and Sylvain Gelly. 2019.
\newblock Parameter-efficient transfer learning for nlp.
\newblock In \emph{ICML}.

\bibitem[{Hu et~al.(2022)Hu, Shen, Wallis, Allen-Zhu, Li, Wang, Wang, and Chen}]{hu2021lora}
Edward~J Hu, Yelong Shen, Phillip Wallis, Zeyuan Allen-Zhu, Yuanzhi Li, Shean Wang, Lu~Wang, and Weizhu Chen. 2022.
\newblock Lora: Low-rank adaptation of large language models.
\newblock In \emph{ICLR}.

\bibitem[{Izmailov et~al.(2018)Izmailov, Podoprikhin, Garipov, Vetrov, and Wilson}]{izmailov_averaging_2018}
Pavel Izmailov, Dmitrii Podoprikhin, Timur Garipov, Dmitry Vetrov, and Andrew~Gordon Wilson. 2018.
\newblock Averaging {Weights} {Leads} to {Wider} {Optima} and {Better} {Generalization}.
\newblock In \emph{Uncertainty in Artificial Intelligence}.

\bibitem[{Jin et~al.(2020)Jin, Jin, Zhou, and Szolovits}]{TextFooler}
Di~Jin, Zhijing Jin, Joey~Tianyi Zhou, and Peter Szolovits. 2020.
\newblock \href {https://aaai.org/ojs/index.php/AAAI/article/view/6311} {Is {BERT} really robust? {A} strong baseline for natural language attack on text classification and entailment}.
\newblock In \emph{AAAI}.

\bibitem[{Li et~al.(2019)Li, Ji, Du, Li, and Wang}]{li2018textbugger}
Jinfeng Li, Shouling Ji, Tianyu Du, Bo~Li, and Ting Wang. 2019.
\newblock \href {https://www.ndss-symposium.org/ndss-paper/textbugger-generating-adversarial-text-against-real-world-applications/} {Textbugger: Generating adversarial text against real-world applications}.
\newblock In \emph{Annual Network and Distributed System Security Symposium}.

\bibitem[{Liu et~al.(2019)Liu, Ott, Goyal, Du, Joshi, Chen, Levy, Lewis, Zettlemoyer, and Stoyanov}]{RoBERTa}
Yinhan Liu, Myle Ott, Naman Goyal, Jingfei Du, Mandar Joshi, Danqi Chen, Omer Levy, Mike Lewis, Luke Zettlemoyer, and Veselin Stoyanov. 2019.
\newblock Roberta: {A} {R}obustly {O}ptimized {BERT} {P}retraining approach.
\newblock In \emph{arXiv}.

\bibitem[{Madry et~al.(2018)Madry, Makelov, Schmidt, Tsipras, and Vladu}]{madry_towards_2017}
Aleksander Madry, Aleksandar Makelov, Ludwig Schmidt, Dimitris Tsipras, and Adrian Vladu. 2018.
\newblock Towards deep learning models resistant to adversarial attacks.
\newblock In \emph{ICLR}.

\bibitem[{Miyato et~al.(2018)Miyato, Maeda, Koyama, and Ishii}]{miyato2018virtual}
Takeru Miyato, Shin-ichi Maeda, Masanori Koyama, and Shin Ishii. 2018.
\newblock Virtual adversarial training: a regularization method for supervised and semi-supervised learning.
\newblock In \emph{IEEE transactions on pattern analysis and machine intelligence}.

\bibitem[{Neyshabur et~al.(2020)Neyshabur, Sedghi, and Zhang}]{neyshabur2020being}
Behnam Neyshabur, Hanie Sedghi, and Chiyuan Zhang. 2020.
\newblock What is being transferred in transfer learning?
\newblock In \emph{NeurIPS}.

\bibitem[{Nguyen and Le(2024)}]{nguyen2024generalizability}
Tuc Nguyen and Thai Le. 2024.
\newblock Generalizability of mixture of domain-specific adapters from the lens of signed weight directions and its application to effective model pruning.
\newblock In \emph{ACL}.

\bibitem[{Pfeiffer et~al.(2020)Pfeiffer, Kamath, R{\"u}ckl{\'e}, Cho, and Gurevych}]{pfeiffer2020adapterfusion}
Jonas Pfeiffer, Aishwarya Kamath, Andreas R{\"u}ckl{\'e}, Kyunghyun Cho, and Iryna Gurevych. 2020.
\newblock Adapterfusion: Non-destructive task composition for transfer learning.
\newblock In \emph{arXiv}.

\bibitem[{Ren et~al.(2019)Ren, Deng, He, and Che}]{PWWS}
Shuhuai Ren, Yihe Deng, Kun He, and Wanxiang Che. 2019.
\newblock \href {https://www.aclweb.org/anthology/P19-1103} {{Generating Natural Language Adversarial Examples through Probability Weighted Word Saliency}}.
\newblock In \emph{ACL}.

\bibitem[{Si et~al.(2021)Si, Zhang, Qi, Liu, Wang, Liu, and Sun}]{si2021better}
Chenglei Si, Zhengyan Zhang, Fanchao Qi, Zhiyuan Liu, Yasheng Wang, Qun Liu, and Maosong Sun. 2021.
\newblock Better robustness by more coverage: Adversarial and mixup data augmentation for robust finetuning.
\newblock In \emph{ACL-IJCNLP}.

\bibitem[{Wang et~al.(2019)Wang, Singh, Michael, Hill, Levy, and Bowman}]{wang2018glue}
Alex Wang, Amanpreet Singh, Julian Michael, Felix Hill, Omer Levy, and Samuel~R Bowman. 2019.
\newblock Glue: A multi-task benchmark and analysis platform for natural language understanding.
\newblock In \emph{ICLR}.

\bibitem[{Wang et~al.(2022)Wang, Agarwal, Mukherjee, Liu, Gao, Awadallah, and Gao}]{wang-etal-2022-adamix}
Yaqing Wang, Sahaj Agarwal, Subhabrata Mukherjee, Xiaodong Liu, Jing Gao, Ahmed~Hassan Awadallah, and Jianfeng Gao. 2022.
\newblock {A}da{M}ix: Mixture-of-adaptations for parameter-efficient model tuning.
\newblock In \emph{EMNLP}.

\bibitem[{Wortsman et~al.(2022)Wortsman, Ilharco, Gadre, Roelofs, Gontijo-Lopes, Morcos, Namkoong, Farhadi, Carmon, Kornblith et~al.}]{wortsman2022model}
Mitchell Wortsman, Gabriel Ilharco, Samir~Ya Gadre, Rebecca Roelofs, Raphael Gontijo-Lopes, Ari~S Morcos, Hongseok Namkoong, Ali Farhadi, Yair Carmon, Simon Kornblith, et~al. 2022.
\newblock Model soups: averaging weights of multiple fine-tuned models improves accuracy without increasing inference time.
\newblock In \emph{ICML}.

\bibitem[{Xie et~al.(2019)Xie, Tan, Gong, Wang, Yuille, and Le}]{xie_adversarial_2019}
Cihang Xie, Mingxing Tan, Boqing Gong, Jiang Wang, Alan Yuille, and Quoc~V Le. 2019.
\newblock Adversarial examples improve image recognition.
\newblock In \emph{arXiv}.

\bibitem[{Xu et~al.(2021)Xu, Liu, Li, Jain, and Tang}]{xu2021robust}
Han Xu, Xiaorui Liu, Yaxin Li, Anil Jain, and Jiliang Tang. 2021.
\newblock To be robust or to be fair: Towards fairness in adversarial training.
\newblock In \emph{ICML}.

\bibitem[{Zang et~al.(2020)Zang, Qi, Yang, Liu, Zhang, Liu, and Sun}]{zang2019word}
Yuan Zang, Fanchao Qi, Chenghao Yang, Zhiyuan Liu, Meng Zhang, Qun Liu, and Maosong Sun. 2020.
\newblock Word-level textual adversarial attacking as combinatorial optimization.
\newblock In \emph{ACL}.

\bibitem[{Zhang et~al.(2018{\natexlab{a}})Zhang, Ciss{\'{e}}, Dauphin, and Lopez{-}Paz}]{mixup}
Hongyi Zhang, Moustapha Ciss{\'{e}}, Yann~N. Dauphin, and David Lopez{-}Paz. 2018{\natexlab{a}}.
\newblock \href {https://openreview.net/forum?id=r1Ddp1-Rb} {mixup: Beyond empirical risk minimization}.
\newblock In \emph{ICLR}.

\bibitem[{Zhang et~al.(2018{\natexlab{b}})Zhang, Cisse, Dauphin, and Lopez-Paz}]{zhang2017mixup}
Hongyi Zhang, Moustapha Cisse, Yann~N Dauphin, and David Lopez-Paz. 2018{\natexlab{b}}.
\newblock mixup: Beyond empirical risk minimization.
\newblock In \emph{ICLR}.

\bibitem[{Zhu et~al.(2020)Zhu, Cheng, Gan, Sun, Goldstein, and Liu}]{zhu2019freelb}
Chen Zhu, Yu~Cheng, Zhe Gan, Siqi Sun, Tom Goldstein, and Jingjing Liu. 2020.
\newblock Freelb: Enhanced adversarial training for natural language understanding.
\newblock In \emph{ICLR}.

\end{thebibliography}

\newpage
\pagebreak
\newpage
\pagebreak

\clearpage
\appendix
\section{Appendix}
\label{sec:appendix}

\subsection{Dataset statistics}
\label{data_length_statistic}

Table~\ref{dataset_statistics} shows the number of instances for each dataset divided by training and test set and linguistic statistics over 5 evaluation datasets.

\renewcommand{\tabcolsep}{1.75pt}
\begin{table*}[t]
\setlength{\belowcaptionskip}{-0.1cm}
\footnotesize
\centering
\begin{tabular}{cccccc}
\toprule
     \textbf{Data Source} & \# \textbf{Training Example} & \# \textbf{Test Example} & \textbf{\begin{tabular}[c]{@{}c@{}}Average \\ Document Length\end{tabular}} & \textbf{\begin{tabular}[c]{@{}c@{}}Average \\ Sentence Length\end{tabular}} & \textbf{\begin{tabular}[c]{@{}c@{}}Average \\ $\#$ Sentences per Document\end{tabular}} \\
     \midrule
     \textbf{MRPC} & 3,668 & 408 &21.9 & 21.1  &  1.0  \\
     \textbf{QNLI} & 104,743& 5,463&18.2  & 18.0 &  1.0  \\
     \textbf{RTE} & 2,490 & 277 &26.2 & 18.1 &  1.4  \\
     \textbf{SST} & 67,347 & 872 &10.4 & 10.4 &  1.0  \\
     \textbf{IMDB} & 22,500 & 2,500 & 233.8 & 21.6 &  10.8  \\
     \hline
\end{tabular}
\caption{Number of instances for each dataset divided by training and test set and linguistic statistics.} 
\label{dataset_statistics}
\end{table*}

\subsection{Training details} \label{sec:transfer_learning}
Tables \ref{fully_finetuning},  \ref{adapter_finetuning} show detailed hyper-parameter in our experiments.
\begin{table*}
\small
\begin{center}
    \begin{tabular}{@{\hskip1pt}l@{\hskip1pt}|@{\hskip1pt}c@{\hskip1pt}|c@{\hskip1pt}|c@{\hskip1pt}|c@{\hskip1pt}|c@{\hskip1pt}|@{\hskip1pt}c @{\hskip1pt}|@{\hskip1pt}}
        \toprule \bf Task & Learning rate & epoch   & train batch size        & evaluation batch size             \\ \midrule \bottomrule

\multicolumn{5}{c}{\textbf{BERT\textsubscript{BASE}}}\\ \midrule	
       MRPC& 2e-5 & 3 & 16 & 8 \\
       QNLI & 3e-5 & 5 & 32 & 8  \\
       RTE & 2e-5 & 3 & 16 & 8 \\
       SST2 & 2e-5 & 3 & 32 & 8 \\
       IMDB & 5e-5 & 5 & 16 & 8 \\
       \midrule
    \multicolumn{5}{c}{\textbf{RoBERTa\textsubscript{LARGE}}}\\ \midrule	
       MRPC & 3e-5 & 10 & 32 & 16\\
       QNLI & 2e-4 & 5 & 32 & 16\\
       RTE & 3e-5 & 10 & 32 & 16 \\
       SST2 & 2e-5 & 5 & 32 & 16  \\
       IMDB & 5e-5 & 5 & 32 & 16  \\
    \bottomrule
\end{tabular}
\end{center}
\caption{Hyperparameter configurations for fully finetuning on various tasks.}
\label{fully_finetuning}
\end{table*}

\begin{table*}
\small
\begin{center}
    \begin{tabular}{@{\hskip1pt}l@{\hskip1pt}|@{\hskip1pt}c@{\hskip1pt}|c@{\hskip1pt}|c@{\hskip1pt}|c@{\hskip1pt}|c@{\hskip1pt}|@{\hskip1pt}c @{\hskip1pt}|@{\hskip1pt}}
        \toprule \bf Task & Learning rate & epoch       &batch size        &warmup                &weight decay  & adapter size   \\ \midrule \bottomrule

\multicolumn{7}{c}{\textbf{BERT\textsubscript{BASE}}}\\ \midrule	
       MRPC& 4e-4 & 5 & 32 & 0.06 & 0.1 & 256 \\
       QNLI&4e-4&20&32& 0.06 & 0.1 & 256 \\
       RTE&4e-4&5&32& 0.06 & 0.1 & 256 \\
       SST2&4e-4&10&32& 0.06 & 0.1 & 256 \\
       IMDB&4e-4&5&32& 0.06 & 0.1 & 256 \\
       \midrule
    \multicolumn{7}{c}{\textbf{RoBERTa\textsubscript{LARGE}}}\\ \midrule	
       MRPC&3e-4&5&64&0.6&0.1&64\\
       QNLI&3e-4&20&64&0.6&0.1&64\\
       RTE&3e-4&5&64&0.6&0.1&64\\
       SST2&3e-4&10&64& 0.6 & 0.1 & 64 \\
       IMDB&3e-4&5&64& 0.6 & 0.1 & 64 \\
    \bottomrule
\end{tabular}
\end{center}
\caption{Hyperparameter configurations for adapter finetuning on various tasks.}
\label{adapter_finetuning}
\end{table*}

\subsection{TextAttack Configuration}
\label{textattack_configuration}
For the \textit{TextFooler} attack, we set the minimum embedding cosine similarity between a word and its synonyms as 0.85, and the minimum Universal Sentence Encoder (USE) similarity is 0.84.
For the \textit{BAE} word-based attack, we set the threshold for cosine similarity of USE as 0.94, and the window size is 15.
For the \textit{PS} we set the maximum number of iteration times to 10 and the population
size to 60.
For the \textit{DeepWordBug}, we set the maximum difference in edit distance to a constant 30 for each sample.
For the \textit{TextBugger}, we set top-5 nearest neighbors in a context-aware word vector space, and the semantic similarity threshold for USE is set as 0.8.

\subsection{Detailed Evaluation Results}
\label{detail_result_known_attack}
Table \ref{full_clean_adv_traditionadv_training}, \ref{full_clean_adv_traditionadv_training_bert} show detailed results on the generalization and adversarial robustness of RoBERTa, BERT.

\renewcommand{\tabcolsep}{3pt}
\begin{table*}[htb!]
\footnotesize
\begin{tabular}{lccccccccccccccc}
\toprule
\multicolumn{1}{c}{\multirow{3}{*}{\textit{\textbf{Methods}}}} & \multicolumn{2}{c}{MRPC}  & \multicolumn{2}{c}{{QNLI}}    & \multicolumn{2}{c}{RTE}    & \multicolumn{2}{c}{{SST2}} & \multicolumn{2}{c}{IMDB}           \\
\cmidrule(lr){2-3} \cmidrule(lr){4-5} \cmidrule(lr){6-7} \cmidrule(lr){8-9} \cmidrule(lr){10-11}
\multicolumn{1}{c}{}  & \multicolumn{1}{c}{Clean} & \multicolumn{1}{c}{Attack}&  \multicolumn{1}{c}{Clean} & \multicolumn{1}{c}{Attack}  & \multicolumn{1}{c}{Clean} & \multicolumn{1}{c}{Attack}  & \multicolumn{1}{c}{Clean} & \multicolumn{1}{c}{Attack}  & \multicolumn{1}{c}{Clean} & \multicolumn{1}{c}{Attack}  \\  
\cmidrule(lr){1-11}

\multicolumn{11}{l}{\textit{\textbf{TextFooler}}} \\
\multicolumn{1}{l}{\textit{CleanOnly}}                                 & \multicolumn{1}{c}{90.0}  & \multicolumn{1}{c}{51.1}  & \multicolumn{1}{c}{94.8}  & \multicolumn{1}{c}{56.0} & \multicolumn{1}{c}{85.2}  & \multicolumn{1}{c}{33.6} & \multicolumn{1}{c}{95.9}  & \multicolumn{1}{c}{42.4} & \multicolumn{1}{c}{92.5}  & 50.8  &\\ 

\multicolumn{1}{l}{\textit{AdvOnly}}                           & \multicolumn{1}{c}{68.4}  & \multicolumn{1}{c}{68.6}  & \multicolumn{1}{c}{49.4}  & \multicolumn{1}{c}{66.4} & \multicolumn{1}{c}{52.7}  & \multicolumn{1}{c}{76.2} & \multicolumn{1}{c}{51.0}  & \multicolumn{1}{c}{59.0}  & \multicolumn{1}{c}{51.1}  & 76.9   & \\ 

\multicolumn{1}{l}{\textit{AdvTrain}}                       & \multicolumn{1}{c}{87.8}  & \multicolumn{1}{c}{64.1} & \multicolumn{1}{c}{92.0}  & \multicolumn{1}{c}{64.4} &  \multicolumn{1}{c}{84.5}  & \multicolumn{1}{c}{62.7}  &  \multicolumn{1}{c}{95.2}  & \multicolumn{1}{c}{56.5} &  \multicolumn{1}{c}{90.6}  & 75.7  & \\ 

\multicolumn{1}{l}{\textit{ModelSoup}}                                  & \multicolumn{1}{c}{{87.3}}  & \multicolumn{1}{c}{{53.9}} &  \multicolumn{1}{c}{{67.0}}  & \multicolumn{1}{c}{{65.0}} &  \multicolumn{1}{c}{{85.1}}  & \multicolumn{1}{c}{{56.8}} &  \multicolumn{1}{c}{{94.8}}  & \multicolumn{1}{c}{{51.5}} &  \multicolumn{1}{c}{{89.2}}  & {67.7} &  \\ 

\multicolumn{1}{l}{{\mymethod}}          & \textbf{89.9}  &      \textbf{68.6}               & \textbf{94.4} & \textbf{66.7}    & \textbf{85.0} & \textbf{76.6}   &\textbf{96.3}& \textbf{58.9}   & \textbf{92.6} & \textbf{76.8}   \\ 
\cmidrule(lr){1-11}

\multicolumn{11}{l}{\textbf{PW}} \\ 
\multicolumn{1}{l}{\textit{CleanOnly}}                                 & \multicolumn{1}{c}{90.0}  & \multicolumn{1}{c}{60.3}   & \multicolumn{1}{c}{94.8}  & \multicolumn{1}{c}{63.2}  & \multicolumn{1}{c}{85.2}   & \multicolumn{1}{c}{35.4}   & \multicolumn{1}{c}{95.9}   & \multicolumn{1}{c}{67.7}  & \multicolumn{1}{c}{92.5}  & 54.1   & \\ 

\multicolumn{1}{l}{\textit{AdvOnly}}                           & \multicolumn{1}{c}{68.2}  & \multicolumn{1}{c}{67.4}   & \multicolumn{1}{c}{52.9}  & \multicolumn{1}{c}{88.6}   & \multicolumn{1}{c}{51.3}  & \multicolumn{1}{c}{72.6}   & \multicolumn{1}{c}{50.9}  & \multicolumn{1}{c}{74.6}   & \multicolumn{1}{c}{54.6}  & 90.8  & \\ 

\multicolumn{1}{l}{\textit{AdvTrain}}                       & \multicolumn{1}{c}{87.3}  & \multicolumn{1}{c}{66.8}   & \multicolumn{1}{c}{92.2}  & \multicolumn{1}{c}{79.1}   & \multicolumn{1}{c}{82.9}  & \multicolumn{1}{c}{63.7}   & \multicolumn{1}{c}{95.3}  & \multicolumn{1}{c}{74.2}   & \multicolumn{1}{c}{89.1}  & 68.4   & \\

\multicolumn{1}{l}{\textit{ModelSoup}}                                  & \multicolumn{1}{c}{{71.1}}  & \multicolumn{1}{c}{{66.1}}   & \multicolumn{1}{c}{{93.7}}  & \multicolumn{1}{c}{{74.8}}  & \multicolumn{1}{c}{{62.5}}  & \multicolumn{1}{c}{{62.1}}  & \multicolumn{1}{c}{{95.4}}  & \multicolumn{1}{c}{{51.5}}   & \multicolumn{1}{c}{{89.6}}  & {60.7}  & \\ 

\multicolumn{1}{l}{{\mymethod}}  & \textbf{89.3}&   \textbf{67.9} & \textbf{94.7}& \textbf{82.4} & \textbf{85.1}& \textbf{71.9} & \textbf{95.8}& \textbf{74.7} & \textbf{92.5}& \textbf{84.3} \\

\cmidrule(lr){1-11} 

\multicolumn{11}{l}{\textbf{BAE}} \\ 
\multicolumn{1}{l}{\textit{CleanOnly}}    & 90.0    & 55.5 &  94.8 & 50.9 &  85.2 & 39.4 &  95.9 & 29.3 &  92.5 & 52.9                         \\ 

\multicolumn{1}{l}{\textit{AdvOnly}}  & 68.4 & 65.8  &  51.5 & 60.4 &  51.3 & 69.1 &  40.7 & 67.2  & 53.0 & 91.2 \\ 

\multicolumn{1}{l}{\textit{AdvTrain}}   & 88.7 & 60.1 & 94.1 & 58.0 & 79.1 & 62.0 & 94.2 & 55.2 & 91.3 & 70.2   \\

\multicolumn{1}{l}{\textit{ModelSoup}}  & {72.1} & {62.3} & {88.8} & {56.6}  & {75.3} & {61.4}  & {90.1} & {51.3} & {86.9} & {65.3}\\ 

\multicolumn{1}{l}{{\mymethod}} & \textbf{89.4} &\textbf{65.1}  & \textbf{94.5} & \textbf{61.6} & \textbf{84.9} & \textbf{68.3}  & \textbf{96.4} & \textbf{66.8}  & \textbf{92.9} & \textbf{86.3} \\

\cmidrule(lr){1-11}
\multicolumn{11}{l}{\textbf{PS}} \\ 
\multicolumn{1}{l}{\textit{CleanOnly}} & 90.0 & 57.2  & 94.8 & 60.6  & 85.2 & 37.8  & 95.9 & 42.4  & 92.5 & 53.4        \\ 

\multicolumn{1}{l}{\textit{AdvOnly}} & 68.4 & 63.2 & 77.2 & 67.2 & 52.3 & 70.4 & 49.7 & 69.0 & 50.2 & 90.9\\ 

\multicolumn{1}{l}{\textit{AdvTrain}} & 88.3 & 60.4 & 93.3 & 65.5 & 81.6 & 66.7 & 96.6 & 65.5 & 90.8 & 72.3    \\

\multicolumn{1}{l}{\textit{ModelSoup}} & {70.1} & {56.2} & {87.2} & {60.9} & {58.5} & {62.2} & {85.1} & {41.5}   & {82.4} & {67.2}   \\ 

\multicolumn{1}{l}{{\mymethod}}  & \textbf{89.3} & \textbf{62.9} & \textbf{94.6} & \textbf{68.2} & \textbf{85.8} & \textbf{69.5} & \textbf{96.6} & \textbf{68.9}  & \textbf{92.8} & \textbf{81.3} \\

\cmidrule(lr){1-11}
\multicolumn{11}{l}{\textbf{DeepWordBug}} \\ 
\multicolumn{1}{l}{\textit{CleanOnly}}       & 90.0 & 66.3     & 94.8  & 65.8  & 85.2 &   52.7   & 95.9 & 52.6  & 92.5 & 55.1             \\ 

\multicolumn{1}{l}{\textit{AdvOnly}} & 66.4 & 76.3 & 55.4 & 73.1 & 51.6 & 73.6 & 39.9 & 79.9 & 50.0 & 93.2  \\ 

\multicolumn{1}{l}{\textit{AdvTrain}} & 88.5 & 70.2 & 93.3 & 68.0 &  79.4 & 66.7 & 93.9 & 73.5 &91.7 & 78.6  \\

\multicolumn{1}{l}{\textit{ModelSoup}}  & {68.4} & {70.0} & {61.0} & {64.8} & {55.6} & {67.7} & {84.6} & {62.3} &{78.0} & {69.6}   \\ 

\multicolumn{1}{l}{{\mymethod}}  & \textbf{90.6} & \textbf{75.9} &  \textbf{94.7} & \textbf{71.3} & \textbf{84.9} & \textbf{71.9} &  \textbf{96.5} & \textbf{78.9} &  \textbf{92.4} & \textbf{84.2}\\

\cmidrule(lr){1-11}
\multicolumn{11}{l}{\textbf{TextBugger}} \\ 
\multicolumn{1}{l}{\textit{CleanOnly}} & 90.0 & 65.8        & 94.8 &    65.5    &     85.2 & 47.3  & 95.9 & 60.3 & 92.5 & 61.0          \\ 

\multicolumn{1}{l}{\textit{AdvOnly}}  & 62.5 & 79.2 & 50.2 & 79.1 & 52.7 & 71.9 & 50.9 & 71.4 & 51.2 & 89.0\\ 

\multicolumn{1}{l}{\textit{AdvTrain}} & 88.2 & 73.2 & 94.5 & 72.2 & 80.9 & 67.3  & 90.7 & 69.6  & 90.9 & 76.3 \\

\multicolumn{1}{l}{\textit{ModelSoup}} & {77.2} & {55.7} & {56.4} & {73.2} & {61.0} & {58.9} & {75.2} & {55.0}  &{74.6} & {70.2}  \\ 

\multicolumn{1}{l}{{\mymethod}}  & \textbf{89.7} & \textbf{77.5} & \textbf{95.2} & \textbf{78.6} & \textbf{85.0} & \textbf{70.9} & \textbf{96.3} & \textbf{70.9} & \textbf{92.4} &\textbf{85.4} \\

\bottomrule
\end{tabular}
\caption{Model performance of independent clean and adversarial training, traditional adversarial training with RoBERTa under TextFooler, and PW textual attack. In \textbf{bold} means better performance compared to Adversarial training. In {red} means performance is worse than Adversarial training.}
\label{full_clean_adv_traditionadv_training}
\end{table*}

\renewcommand{\tabcolsep}{3pt}
\begin{table*}[htb!]
\footnotesize
\begin{tabular}{lccccccccccccccc}
\toprule
\multicolumn{1}{c}{\multirow{3}{*}{\textit{\textbf{Methods}}}} & \multicolumn{2}{c}{MRPC}  & \multicolumn{2}{c}{{QNLI}}    & \multicolumn{2}{c}{RTE}    & \multicolumn{2}{c}{{SST2}} & \multicolumn{2}{c}{IMDB}           \\
\cmidrule(lr){2-3} \cmidrule(lr){4-5} \cmidrule(lr){6-7} \cmidrule(lr){8-9} \cmidrule(lr){10-11}
\multicolumn{1}{c}{}  & \multicolumn{1}{c}{Clean} & \multicolumn{1}{c}{Attack}&  \multicolumn{1}{c}{Clean} & \multicolumn{1}{c}{Attack}  & \multicolumn{1}{c}{Clean} & \multicolumn{1}{c}{Attack}  & \multicolumn{1}{c}{Clean} & \multicolumn{1}{c}{Attack}  & \multicolumn{1}{c}{Clean} & \multicolumn{1}{c}{Attack}  \\  
\cmidrule(lr){1-11}

\multicolumn{11}{l}{\textit{\textbf{TextFooler}}} \\
\multicolumn{1}{l}{\textit{CleanOnly}}   & 83.3  & 64.8 & 90.5 & 62.9 & 65.0 & 35.1 & 92.5 & 52.5 & 88.9 & 52.0\\ 

\multicolumn{1}{l}{\textit{AdvOnly}}    & 35.6 &66.9   &75.3 & 88.9 & 46.9& 58.7   & 23.6 & 68.6   & 60.1 & 87.3      \\ 

\multicolumn{1}{l}{\textit{AdvTrain}} & 78.7 & 68.9 & 86.8 & 73.2 & 62.1 & 46.6 & 92.5 & 67.9 & 86.7 & 55.5    \\
\multicolumn{1}{l}{\textit{ModelSoup}}  & {68.5} & {57.9} & {76.6} & {62.4} & {54.5} & {40.9} & {83.2} & {55.9} & {68.9} & {59.4} \\ 

\multicolumn{1}{l}{{\mymethod}}  & \textbf{82.1} & \textbf{68.0} & \textbf{89.6}& \textbf{84.2} & \textbf{64.1} & \textbf{55.3} & \textbf{92.3} & \textbf{67.9} & \textbf{87.9} & \textbf{83.2}    \\ 

\cmidrule(lr){1-11}

\multicolumn{11}{l}{\textbf{PW}} \\ 
\multicolumn{1}{l}{\textit{CleanOnly}}  & 83.3 &60.5& 90.5 & 59.8 & 65.0 & 39.5& 92.5 & 40.2 & 88.9 & 55.6  \\ 

\multicolumn{1}{l}{\textit{AdvOnly}}  & 54.7 & 74.4 & 76.6& 93.2 & 46.2 & 56.7 & 19.3 & 82.6 & 58.5 & 84.3\\ 

\multicolumn{1}{l}{\textit{AdvTrain}}  & 78.9 & 70.9 & 85.3 & 73.3 & 59.9 & 50.4 & 90.5 & 72.3 & 87.9 & 65.6    \\

\multicolumn{1}{l}{\textit{ModelSoup}}     & {65.4} & {68.3} & {79.8} & {73.2} & {55.3} & {51.6} & {81.7} & {63.4} & {68.4} & {57.2}         \\ 

\multicolumn{1}{l}{{\mymethod}}  & \textbf{81.2} & \textbf{72.0} & \textbf{88.3} & \textbf{84.5}  & \textbf{63.6} & \textbf{55.3} & \textbf{91.7} & \textbf{77.1}  & \textbf{88.2} & \textbf{73.6}\\

\cmidrule(lr){1-11} 

\multicolumn{11}{l}{\textbf{BAE}} \\ 
\multicolumn{1}{l}{\textit{CleanOnly}}   & 83.3 & 60.8& 90.5 & 53.1 & 65.0 &37.9& 92.5 & 27.7 & 88.9  & 49.6        \\ 

\multicolumn{1}{l}{\textit{AdvOnly}}  & 78.7 & 65.8 & 87.6 & 59.5 & 59.6 & 56.7 & 88.0 & 35.5 & 55.6 & 81.4\\ 

\multicolumn{1}{l}{\textit{AdvTrain}} & 79.7 & 66.4 & 90.2 & 56.8 & 63.2 & 53.6 & 91.7 & 35.1 & 88.7 & 60.1    \\

\multicolumn{1}{l}{\textit{ModelSoup}} & {71.7} & {60.1} & {68.8} & {40.1} & {53.1} & {40.2} & {86.4} & {31.3} & {53.6} & {50.8}\\ 

\multicolumn{1}{l}{{\mymethod}} & \textbf{82.0} & \textbf{67.3} & \textbf{90.2} & \textbf{58.2} & \textbf{64.6} & \textbf{55.1} & \textbf{92.1} & \textbf{38.5} & \textbf{88.6} & \textbf{72.8} \\

\cmidrule(lr){1-11}
\multicolumn{11}{l}{\textbf{PS}} \\ 
\multicolumn{1}{l}{\textit{CleanOnly}}   & 83.3 & 56.6 & 90.5 & 64.9 & 65.0 & 33.9 & 92.5 & 40.4& 88.9 & 51.3          \\ 

\multicolumn{1}{l}{\textit{AdvOnly}}  & 76.5 & 78.2 & 79.2 & 64.4 & 60.6 & 44.8 & 86.4 & 57.1 & 55.8 & 78.9  \\ 

\multicolumn{1}{l}{\textit{AdvTrain}} & 80.4 & 71.9 & 89.5 & 67.4 & 63.5 & 41.4 & 93.0 & 59.3 &  85.8 & 65.2  \\

\multicolumn{1}{l}{\textit{ModelSoup}} & {71.0} & {68.6} & {83.3} & {59.4} & {52.7} & {37.2} & {85.6} & {48.7} & {72.3} &  {61.3}\\ 

\multicolumn{1}{l}{{\mymethod}} & \textbf{81.3} & \textbf{75.2} & \textbf{90.4} & \textbf{65.3} & \textbf{64.7} & \textbf{43.5} & \textbf{92.1} & \textbf{57.1}  & \textbf{87.8} & \textbf{73.5}\\

\cmidrule(lr){1-11}
\multicolumn{11}{l}{\textbf{DeepWordBug}} \\ 
\multicolumn{1}{l}{\textit{CleanOnly}}   & 83.3 &51.3 & 90.5 & 52.5& 65.0 & 33.6& 92.5 & 40.3 & 88.9 & 52.2    \\ 

\multicolumn{1}{l}{\textit{AdvOnly}}  & 75.2 & 75.4 & 50.8 & 71.8 & 54.5 & 66.4 & 47.4 & 67.0  & 58.9 & 75.3\\ 

\multicolumn{1}{l}{\textit{AdvTrain}}  & 81.6 & 67.7 & 90.6 & 62.9 & 61.4 & 56.3 & 91.0 & 63.5 & 87.2 & 71.1 \\

\multicolumn{1}{l}{\textit{ModelSoup}} & {68.9} & {53.0} & {76.9} & {53.0} & {54.5} & 58.4 & {86.4} & {48.3} & {52.6} & {60.9}  \\ 

\multicolumn{1}{l}{{\mymethod}} & \textbf{82.4} & \textbf{72.5} & {89.3} & \textbf{68.2} & \textbf{63.1} & \textbf{64.3} & \textbf{91.4} & \textbf{66.0} & \textbf{88.0} & \textbf{74.1}\\

\cmidrule(lr){1-11}
\multicolumn{11}{l}{\textbf{TextBugger}} \\ 
\multicolumn{1}{l}{\textit{CleanOnly}} & 83.3 & 66.0 & 90.5 & 62.0 & 65.0 & 37.8& 92.5 & 50.9  & 88.9   & 59.3    \\ 

\multicolumn{1}{l}{\textit{AdvOnly}} & 38.7 & 76.0 & 52.1 & 69.0 & 45.1 & 65.7 & 23.2 & 76.6 & 53.9 & 88.8\\ 

\multicolumn{1}{l}{\textit{AdvTrain}} & 80.6 & 73.3 & 90.4 & 67.8 & 57.8 & 54.2 & 93.5 & 70.0 & 88.7 & 82.3  \\

\multicolumn{1}{l}{\textit{ModelSoup}} & {70.3} & {66.0} & {81.5} & {59.4} & {54.5} & 55.3 & {74.9} & {46.6} & {53.3} & {73.2} \\ 

\multicolumn{1}{l}{{\mymethod}} & \textbf{82.0} & \textbf{74.7} & \textbf{90.1} & \textbf{68.0} & \textbf{63.5} & \textbf{63.3} & \textbf{92.6} & \textbf{74.1} & \textbf{88.5} & \textbf{84.4} \\

\bottomrule
\end{tabular}
\caption{Model performance of independent clean and adversarial training, traditional adversarial training with BERT under TextFooler, and PW textual attack. In \textbf{bold} means better performance compared to Adversarial training. In {red} means performance is worse than Adversarial training.}
\label{full_clean_adv_traditionadv_training_bert}
\end{table*}

\subsection{Detailed Cross Attack Evaluation}
\label{detailed_cross_attack_m_1}

\paragraph{Average cross-attack evaluation.} Table \ref{avg_cross_attack_word_to_character} and \ref{avg_cross_attack_character_to_word} show average cross-attack evaluation from word-base to character-base and vice versa.
\renewcommand{\tabcolsep}{1.5pt}
\begin{table*}[htb!]
\footnotesize
\begin{tabular}{lccc|cc|cc|cc|cc|cc|cc|cc}
\toprule
&\multicolumn{1}{c}{\textit{\textbf{Attacker}}} & \multicolumn{2}{c}{TF$\rightarrow$DW} & \multicolumn{2}{c}{TF$\rightarrow$TB}  & \multicolumn{2}{c}{BA$\rightarrow$DW} & \multicolumn{2}{c}{BA$\rightarrow$TB} & \multicolumn{2}{c}{PS$\rightarrow$DW} & \multicolumn{2}{c}{PS$\rightarrow$TB}  & \multicolumn{2}{c}{PW$\rightarrow$DW} & \multicolumn{2}{c}{PW$\rightarrow$TB} \\
\cmidrule(lr){3-4} \cmidrule(lr){5-6} \cmidrule(lr){7-8} \cmidrule(lr){9-10} \cmidrule(lr){11-12} \cmidrule(lr){13-14}  \cmidrule(lr){15-16} \cmidrule(lr){17-18}

&\multicolumn{1}{c}{\textit{\textbf{Method}}} & \multicolumn{1}{c}{Clean}& \multicolumn{1}{c}{Adv} & \multicolumn{1}{c}{Clean}& \multicolumn{1}{c}{Adv} &\multicolumn{1}{c}{Clean}& \multicolumn{1}{c}{Adv} &\multicolumn{1}{c}{Clean}& \multicolumn{1}{c}{Adv} &\multicolumn{1}{c}{Clean}& \multicolumn{1}{c}{Adv} &\multicolumn{1}{c}{Clean}& \multicolumn{1}{c}{Adv} &\multicolumn{1}{c}{Clean}& \multicolumn{1}{c}{Adv} &\multicolumn{1}{c}{Clean}& \multicolumn{1}{c}{Adv}\\

\cmidrule(lr){1-18}
\multirow{5}{*}{\begin{sideways} \textit{\textbf{RoBERTa}} \end{sideways}}
&\multicolumn{1}{l}{\textit{CleanOnly}} & 91.7 &58.5 & 91.7&60.0 & 91.7& 58.5 & 91.7&60.0 & 91.7&58.5& 91.7& 60.0 &91.7& 58.5 &91.7& 60.0 \\ 

&\multicolumn{1}{l}{\textit{AdvOnly}} & 54.5 & 68.0 &54.5& 67.1 &53.0& 63.3 &53.0& 61.5 & 59.6& 65.7 &59.6& 67.5 &55.6&62.5 &55.6& 69.3\\ 

&\multicolumn{1}{l}{\textit{AdvTrain}}  &90.0 & 64.7 &90.0& 63.2 &89.5&61.4 &89.5& 58.5 & 90.1& 63.0 &90.1& 63.8 &89.4& 60.3 &89.4& 65.8 \\ 

&\multicolumn{1}{l}{\textit{ModelSoup}} & 84.7& {59.7} &84.7& {59.2}&82.6& {57.4} &82.6& {52.6} &76.7& {57.0} &76.7& {58.9}&82.5& {54.6} &82.5& {61.1} \\

&\multicolumn{1}{l}{{\mymethod}} &\textbf{91.6}& \textbf{66.4} &\textbf{91.6}& \textbf{65.9}  &\textbf{91.6}& \textbf{64.0} &\textbf{91.6}& \textbf{62.6} &\textbf{91.8}& \textbf{65.6} &\textbf{91.8}& \textbf{67.0}  &\textbf{91.5}& \textbf{62.0} &\textbf{91.5}& \textbf{69.0}\\
\bottomrule
\end{tabular}
\caption{Cross Attack Evaluation From Word-based to Character-based methods}
\label{avg_cross_attack_word_to_character}
\end{table*}

\renewcommand{\tabcolsep}{1.5pt}
\begin{table*}[htb!]
\footnotesize
\begin{tabular}{lccc|cc|cc|cc|cc|cc|cc|cc}
\toprule
&\multicolumn{1}{c}{\textit{\textbf{Attacker}}} &  \multicolumn{2}{c}{DW$\rightarrow$TF} & \multicolumn{2}{c}{DW$\rightarrow$BAE} & \multicolumn{2}{c}{DW$\rightarrow$PS} & \multicolumn{2}{c}{DW$\rightarrow$PW}  & \multicolumn{2}{c}{TB$\rightarrow$TF} & \multicolumn{2}{c}{TB$\rightarrow$BAE} & \multicolumn{2}{c}{TB$\rightarrow$PS} & \multicolumn{2}{c}{TB$\rightarrow$PW}  \\
\cmidrule(lr){3-4} \cmidrule(lr){5-6} \cmidrule(lr){7-8} \cmidrule(lr){9-10} \cmidrule(lr){11-12} \cmidrule(lr){13-14}  \cmidrule(lr){15-16} \cmidrule(lr){17-18}

&\multicolumn{1}{c}{\textit{\textbf{Method}}} & \multicolumn{1}{c}{Clean}& \multicolumn{1}{c}{Adv} & \multicolumn{1}{c}{Clean}& \multicolumn{1}{c}{Adv} &\multicolumn{1}{c}{Clean}& \multicolumn{1}{c}{Adv} &\multicolumn{1}{c}{Clean}& \multicolumn{1}{c}{Adv} &\multicolumn{1}{c}{Clean}& \multicolumn{1}{c}{Adv} &\multicolumn{1}{c}{Clean}& \multicolumn{1}{c}{Adv} &\multicolumn{1}{c}{Clean}& \multicolumn{1}{c}{Adv} &\multicolumn{1}{c}{Clean}& \multicolumn{1}{c}{Adv}\\

\cmidrule(lr){1-18}
\multirow{5}{*}{\begin{sideways} \textit{\textbf{RoBERTa}} \end{sideways}}
&\multicolumn{1}{l}{\textit{CleanOnly}} & 91.7 &  46.8  &91.7& 45.6  &91.7& 50.3  &91.7& 56.1 & 91.7&  46.8 & 91.7 &45.6 & 91.7& 50.3 &91.7 & 56.1\\ 

&\multicolumn{1}{l}{\textit{AdvOnly}} & 52.7 & 62.7 & 52.7&60.2 & 52.7& 62.3 & 52.7& 66.0 & 53.5 & 61.3 & 53.5& 56.1 & 53.5& 59.3 & 53.5& 63.4\\ 

&\multicolumn{1}{l}{\textit{AdvTrain}}  & 89.4 & 56.4 & 89.4& 51.5 & 89.4& 54.7 & 89.4& 61.4 & 89.0& 56.0 & 89.0& 52.1& 89.0 &57.4 & 89.0& 59.4  \\ 

&\multicolumn{1}{l}{\textit{ModelSoup}} & 69.5 & {53.5} & 69.5& {44.9}& 69.5& {46.2} &69.5 & {44.3} & 68.9& {52.0} & 68.9& {42.0}& 68.9& {46.0} & 68.9& {50.5} \\
\bottomrule
\end{tabular}
\caption{Cross Attack Evaluation From Character-based to Word-based methods}
\label{avg_cross_attack_character_to_word}
\end{table*}

\paragraph{From Word-based to Character-based attack.}
Table~\ref{cross_attack_word_character_roberta}, \ref{cross_attack_word_charater_bert} show detailed cross evaluation from word-based to character-based attack of RoBERTa and BERT.

\renewcommand{\tabcolsep}{0.8pt}
\begin{table*}[htb!]
\footnotesize
\begin{tabular}{lcccccccccccccccccccc}
\toprule
\multicolumn{1}{c}{\multirow{2}{*}{\textit{\textbf{Methods}}}} & \multicolumn{10}{c}{\textbf{DeepWordBug}}   & \multicolumn{10}{c}{\textbf{TextBugger}}    \\

\cmidrule(lr){2-11} \cmidrule(lr){12-21} 
\multicolumn{1}{c}{}                                                    & \multicolumn{2}{c}{MRPC} & \multicolumn{2}{c}{QNLI}& \multicolumn{2}{c}{RTE}  & \multicolumn{2}{c}{SST2} & \multicolumn{2}{c}{IMDB}  & \multicolumn{2}{c}{MRPC} & \multicolumn{2}{c}{QNLI}& \multicolumn{2}{c}{RTE}  & \multicolumn{2}{c}{SST2} & \multicolumn{2}{c}{IMDB} \\

\cmidrule(lr){2-3} \cmidrule(lr){4-5} \cmidrule(lr){6-7} \cmidrule(lr){8-9} \cmidrule(lr){10-11} \cmidrule(lr){12-13} \cmidrule(lr){14-15} \cmidrule(lr){16-17} \cmidrule(lr){18-19} \cmidrule(lr){20-21}

\multicolumn{1}{c}{}  & \multicolumn{1}{c}{Clean} & \multicolumn{1}{c}{Adv} & \multicolumn{1}{c}{Clean} & \multicolumn{1}{c}{Adv}& \multicolumn{1}{c}{Clean} & \multicolumn{1}{c}{Adv}& \multicolumn{1}{c}{Clean} & \multicolumn{1}{c}{Adv}& \multicolumn{1}{c}{Clean} & \multicolumn{1}{c}{Adv}& \multicolumn{1}{c}{Clean} & \multicolumn{1}{c}{Adv}& \multicolumn{1}{c}{Clean} & \multicolumn{1}{c}{Adv}& \multicolumn{1}{c}{Clean} & \multicolumn{1}{c}{Adv}& \multicolumn{1}{c}{Clean} & \multicolumn{1}{c}{Adv}& \multicolumn{1}{c}{Clean} & \multicolumn{1}{c}{Adv} \\

\cmidrule(lr){1-21}

\multicolumn{1}{l}{\textit{CleanOnly}} & 90.0& 66.3 & 94.8&65.8 & 85.2&52.7 & 95.9&52.6 & 92.5&55.1   & 90.0&65.8  & 94.8&65.5 & 85.2&47.3 & 95.9&60.3 & 92.5&61.0      \\ 

\cmidrule(lr){1-21}

\multicolumn{21}{l}{\textit{\textbf{Clean + TextFooler}}} \\

\multicolumn{1}{l}{\textit{AdvOnly}} & 68.4&70.0 & 49.4&66.6 & 52.7&67.7 & 51.0&60.8 & 51.1&74.9   &68.4&72.6 & 49.4&67.0 & 52.7&64.3 & 51.0&56.5 & 51.1&75.3 \\

\multicolumn{1}{l}{\textit{AdvTrain}}  & 87.8&68.5  & 92.0&64.9 & 84.5&66.2 & 95.2&59.2 & 90.6&64.8 & 87.8&70.4  & 92.0&64.0 & 84.5&57.6 & 95.2&53.2 & 90.6&70.9 \\
\bottomrule
\end{tabular}
\caption{Cross Evaluation (from Word-based to Character-based) with RoBERTa}
\label{cross_attack_word_character_roberta}
\end{table*}

\renewcommand{\tabcolsep}{0.8pt}
\begin{table*}[htb!]
\footnotesize
\begin{tabular}{lcccccccccccccccccccc}
\toprule
\multicolumn{1}{c}{\multirow{2}{*}{\textit{\textbf{Methods}}}} & \multicolumn{10}{c}{\textbf{DeepWordBug}}   & \multicolumn{10}{c}{\textbf{TextBugger}}    \\

\cmidrule(lr){2-11} \cmidrule(lr){12-21} 
\multicolumn{1}{c}{}   & \multicolumn{2}{c}{MRPC} & \multicolumn{2}{c}{QNLI}& \multicolumn{2}{c}{RTE}  & \multicolumn{2}{c}{SST2} & \multicolumn{2}{c}{IMDB} & \multicolumn{2}{c}{MRPC} & \multicolumn{2}{c}{QNLI}& \multicolumn{2}{c}{RTE}  & \multicolumn{2}{c}{SST2} & \multicolumn{2}{c}{IMDB} \\
\cmidrule(lr){2-3} \cmidrule(lr){4-5} \cmidrule(lr){6-7} \cmidrule(lr){8-9} \cmidrule(lr){10-11} \cmidrule(lr){12-13} \cmidrule(lr){14-15} \cmidrule(lr){16-17} \cmidrule(lr){18-19} \cmidrule(lr){20-21}

\multicolumn{1}{c}{}  & \multicolumn{1}{c}{Clean} & \multicolumn{1}{c}{Adv} & \multicolumn{1}{c}{Clean} & \multicolumn{1}{c}{Adv}& \multicolumn{1}{c}{Clean} & \multicolumn{1}{c}{Adv}& \multicolumn{1}{c}{Clean} & \multicolumn{1}{c}{Adv}& \multicolumn{1}{c}{Clean} & \multicolumn{1}{c}{Adv}& \multicolumn{1}{c}{Clean} & \multicolumn{1}{c}{Adv}& \multicolumn{1}{c}{Clean} & \multicolumn{1}{c}{Adv}& \multicolumn{1}{c}{Clean} & \multicolumn{1}{c}{Adv}& \multicolumn{1}{c}{Clean} & \multicolumn{1}{c}{Adv}& \multicolumn{1}{c}{Clean} & \multicolumn{1}{c}{Adv} \\
\cmidrule(lr){1-21}

\multicolumn{1}{l}{\textit{CleanOnly}} & 83.3&51.3 & 90.5&52.5 & 65.0&33.6 & 92.5&40.3 & 88.9&52.2 &83.3&66.0 & 90.5&62.0 & 65.0&37.8 & 92.5&50.9 & 88.9&59.3  \\ 

\cmidrule(lr){1-21}

\multicolumn{21}{l}{\textit{\textbf{Clean + TextFooler}}} \\

\multicolumn{1}{l}{\textit{AdvOnly}} & 35.6&78.6 & 75.3&61.4 & 46.9&58.0& 23.6&64.6 & 60.1&59.9&   35.6&75.8 & 75.3&62.3 & 46.9&53.8 & 23.6&60.4 & 60.1&66.9 \\ 

\multicolumn{1}{l}{\textit{AdvTrain}}  & 78.7&74.2 & 86.8&58.4 & 62.1&50.0 & 92.5&48.7 & 86.7&56.7&  78.7&74.1 & 86.8&57.8 & 62.1&51.0 & 92.5&57.9 & 86.7&63.8  \\
\bottomrule
\end{tabular}
\caption{Cross Attack Evaluation (from Word-based to Character-based) with BERT}
\label{cross_attack_word_charater_bert}
\end{table*}

\paragraph{From Character-based to Word-based attack.}
Table~\ref{cross_attack_character_word_roberta_1}, \ref{cross_attack_character_word_roberta_2},
\ref{cross_attack_character_word_bert_1}, \ref{cross_attack_character_word_bert_2} show detailed cross evaluation from character-based to word-based attack of RoBERTa and BERT.

\renewcommand{\tabcolsep}{0.8pt}
\begin{table*}[htb!]
\footnotesize
\begin{tabular}{lcccccccccccccccccccccc}
\toprule
\multicolumn{1}{c}{\multirow{2}{*}{\textit{\textbf{Methods}}}} & \multicolumn{10}{c}{\textbf{TextFooler}}   & \multicolumn{10}{c}{\textbf{BAE}}    \\

\cmidrule(lr){2-21}
\multicolumn{1}{c}{}   & \multicolumn{2}{c}{MRPC} & \multicolumn{2}{c}{QNLI}& \multicolumn{2}{c}{RTE}  & \multicolumn{2}{c}{SST2} & \multicolumn{2}{c}{IMDB} & \multicolumn{2}{c}{MRPC} & \multicolumn{2}{c}{QNLI}& \multicolumn{2}{c}{RTE}  & \multicolumn{2}{c}{SST2} & \multicolumn{2}{c}{IMDB} \\
\cmidrule(lr){2-3} \cmidrule(lr){4-5} \cmidrule(lr){6-7} \cmidrule(lr){8-9} \cmidrule(lr){10-11} \cmidrule(lr){12-13} \cmidrule(lr){14-15} \cmidrule(lr){16-17} \cmidrule(lr){18-19} \cmidrule(lr){20-21}

\multicolumn{1}{c}{}  & \multicolumn{1}{c}{Clean} & \multicolumn{1}{c}{Adv} & \multicolumn{1}{c}{Clean} & \multicolumn{1}{c}{Adv}& \multicolumn{1}{c}{Clean} & \multicolumn{1}{c}{Adv}& \multicolumn{1}{c}{Clean} & \multicolumn{1}{c}{Adv}& \multicolumn{1}{c}{Clean} & \multicolumn{1}{c}{Adv}& \multicolumn{1}{c}{Clean} & \multicolumn{1}{c}{Adv}& \multicolumn{1}{c}{Clean} & \multicolumn{1}{c}{Adv}& \multicolumn{1}{c}{Clean} & \multicolumn{1}{c}{Adv}& \multicolumn{1}{c}{Clean} & \multicolumn{1}{c}{Adv}& \multicolumn{1}{c}{Clean} & \multicolumn{1}{c}{Adv} \\
\cmidrule(lr){1-21}

\multicolumn{1}{l}{\textit{CleanOnly}} &   90.0&51.1 & 94.8&56.0 & 85.2&33.6 & 95.9&42.4 & 92.5&50.8 & 90.0&55.5 & 94.8&50.9 & 85.2&39.4 & 95.9&29.3 & 92.5&52.9  \\ 

\cmidrule(lr){1-21}

\multicolumn{21}{l}{\textit{\textbf{Clean + DeepWordBug}}} \\

\multicolumn{1}{l}{\textit{AdvOnly}} & 66.4&58.6 & 55.4&64.6 & 51.6&75.7 & 39.9&52.5 & 50.0&61.9 &  66.4&65.1 & 55.5&54.5 & 51.6&66.3 & 39.9&55.2 & 50.0&59.7  \\ 

\multicolumn{1}{l}{\textit{AdvTrain}}  & 88.5&57.8 & 93.3&63.5 & 79.4&51.4 & 93.9&51.0 &91.7&58.3 & 88.5&59.6 & 93.3&54.4 & 79.4&55.4 & 93.9&37.9 & 91.7&50.3 \\
\bottomrule
\end{tabular}
\caption{Cross Attack Evaluation (from Character-based to Word-based) with RoBERTa (1)}
\label{cross_attack_character_word_roberta_1}
\end{table*}

\renewcommand{\tabcolsep}{2pt}
\begin{table*}[htb!]
\footnotesize
\begin{tabular}{lcccccccccc}
\toprule
\multicolumn{1}{c}{\multirow{2}{*}{\textit{\textbf{Methods}}}} & \multicolumn{5}{c}{\textbf{PS}}   & \multicolumn{5}{c}{\textbf{PW}}    \\

\cmidrule(lr){2-6} \cmidrule(lr){7-11} 
\multicolumn{1}{c}{}                                                    & \multicolumn{1}{c}{MRPC} & \multicolumn{1}{c}{QNLI}& \multicolumn{1}{c}{RTE}  & \multicolumn{1}{c}{SST2} & \multicolumn{1}{c}{IMDB}  & \multicolumn{1}{c}{MRPC} & \multicolumn{1}{c}{QNLI}& \multicolumn{1}{c}{RTE}  & \multicolumn{1}{c}{SST2} & \multicolumn{1}{c}{IMDB}\\

\cmidrule(lr){1-11}

\multicolumn{1}{l}{\textit{CleanOnly}} &  57.2 & 60.6 & 37.8 & 42.4 & 53.4 & 60.3 & 63.2 & 35.4 & 67.7 & 54.1  \\ 

\cmidrule(lr){1-11}

\multicolumn{11}{l}{\textit{\textbf{Clean + DeepWordBug}}} \\

\multicolumn{1}{l}{\textit{AdvOnly}} & 64.5 & 60.6 & 68.9 & 52.0 & 65.4 & 68.3 & 68.9 & 72.3 & 54.1 & 65.2 \\ 

\multicolumn{1}{l}{\textit{AdvTrain}} & 61.8 & 54.5 & 48.1 & 49.3 & 59.6 & 64.6 & 67.5 & 56.9 & 56.6 & 61.6\\ 

\multicolumn{1}{l}{\textit{ModelSoup}} & {38.8} & {50.3} & {58.1} & {40.2} &{43.6}& {38.1} & {45.7} & {40.3} & {44.7} & {52.5} \\ 

\multicolumn{1}{l}{{\mymethod}} & \textbf{63.9} & \textbf{59.0} & \textbf{68.5} & \textbf{59.6} & \textbf{63.1} & \textbf{68.5} & \textbf{66.3} & \textbf{72.9} & \textbf{56.8} & \textbf{64.0}\\ 

\cmidrule(lr){1-11}

\multicolumn{11}{l}{\textbf{Clean + TextBugger}} \\ 

\multicolumn{1}{l}{\textit{AdvOnly}} & 64.5 & 58.2 & 68.9 & 47.2 & 57.8 & 64.6 & 67.3 & 72.3 & 54.6 & 58.3 \\ 

\multicolumn{1}{l}{\textit{AdvTrain}} & 62.1 & 59.6 & 60.7 & 49.8 & 54.7 & 62.8 & 63.3 & 66.7  & 49.2 & 54.8 \\

\multicolumn{1}{l}{\textit{ModelSoup}} & {50.7} & {40.0} & {58.5} & {42.8} & {42.1} &  {55.6} & {45.5} & {61.0} & {40.7} & {49.6} \\ 

\multicolumn{1}{l}{{\mymethod}} & \textbf{63.2} & \textbf{59.4} & \textbf{69.3} & \textbf{50.8} & \textbf{56.4}& \textbf{66.1} & \textbf{65.4} & \textbf{72.8} & \textbf{54.4} & \textbf{56.9} \\

\bottomrule
\end{tabular}
\caption{Cross Attack Evaluation (from Character-based to Word-based) with RoBERTa (2)}
\label{cross_attack_character_word_roberta_2}
\end{table*}

\renewcommand{\tabcolsep}{0.8pt}
\begin{table*}[htb!]
\footnotesize
\begin{tabular}{lcccccccccccccccccccc}
\toprule
\multicolumn{1}{c}{\multirow{2}{*}{\textit{\textbf{Methods}}}} & \multicolumn{10}{c}{\textbf{TextFooler}}   & \multicolumn{10}{c}{\textbf{BAE}}    \\

\cmidrule(lr){2-11} \cmidrule(lr){12-21}
\multicolumn{1}{c}{}   & \multicolumn{2}{c}{MRPC} & \multicolumn{2}{c}{QNLI}& \multicolumn{2}{c}{RTE}  & \multicolumn{2}{c}{SST2} & \multicolumn{2}{c}{IMDB} & \multicolumn{2}{c}{MRPC} & \multicolumn{2}{c}{QNLI}& \multicolumn{2}{c}{RTE}  & \multicolumn{2}{c}{SST2} & \multicolumn{2}{c}{IMDB} \\
\cmidrule(lr){2-3} \cmidrule(lr){4-5} \cmidrule(lr){6-7} \cmidrule(lr){8-9} \cmidrule(lr){10-11} \cmidrule(lr){12-13} \cmidrule(lr){14-15} \cmidrule(lr){16-17} \cmidrule(lr){18-19} \cmidrule(lr){20-21}

\multicolumn{1}{c}{}  & \multicolumn{1}{c}{Clean} & \multicolumn{1}{c}{Adv} & \multicolumn{1}{c}{Clean} & \multicolumn{1}{c}{Adv}& \multicolumn{1}{c}{Clean} & \multicolumn{1}{c}{Adv}& \multicolumn{1}{c}{Clean} & \multicolumn{1}{c}{Adv}& \multicolumn{1}{c}{Clean} & \multicolumn{1}{c}{Adv}& \multicolumn{1}{c}{Clean} & \multicolumn{1}{c}{Adv}& \multicolumn{1}{c}{Clean} & \multicolumn{1}{c}{Adv}& \multicolumn{1}{c}{Clean} & \multicolumn{1}{c}{Adv}& \multicolumn{1}{c}{Clean} & \multicolumn{1}{c}{Adv}& \multicolumn{1}{c}{Clean} & \multicolumn{1}{c}{Adv} \\
\cmidrule(lr){1-21}

\multicolumn{1}{l}{\textit{CleanOnly}} &  83.3&64.8 & 90.5&62.9 & 65.0&35.1 & 92.5&52.5 & 88.9&52.0 & 83.3&60.8 & 90.5&53.1 & 65.0&37.9 & 92.5&27.7 & 88.9&49.6\\ 

\cmidrule(lr){1-21}

\multicolumn{21}{l}{\textit{\textbf{Clean + DeepWordBug}}} \\

\multicolumn{1}{l}{\textit{AdvOnly}} & 75.2&75.5 & 50.8&57.7 & 54.5&61.5 & 47.4&58.0 & 58.9&63.6 &  75.2&74.5 & 50.8&60.9 & 54.5&60.7 & 47.4&70.0 & 58.9&56.8 \\ 

\multicolumn{1}{l}{\textit{AdvTrain}} & 81.6&67.6 & 90.6&56.1 & 61.4&46.2 & 91.0&54.3 & 87.2&59.5& 81.6&68.5  & 90.6&58.8 & 61.4&47.8 & 91.0&31.9 & 87.2&53.7 \\
\bottomrule
\end{tabular}
\caption{Cross Attack Evaluation (from Character-based to Word-based) with BERT (1)}
\label{cross_attack_character_word_bert_1}
\end{table*}

\renewcommand{\tabcolsep}{2pt}
\begin{table*}[htb!]
\footnotesize
\begin{tabular}{lcccccccccccc}
\toprule
\multicolumn{1}{c}{\multirow{2}{*}{\textit{\textbf{Methods}}}} & \multicolumn{5}{c}{\textbf{PS}}   & \multicolumn{5}{c}{\textbf{PW}}    \\

\cmidrule(lr){2-6} \cmidrule(lr){7-11} 
\multicolumn{1}{c}{}                                                    & \multicolumn{1}{c}{MRPC} & \multicolumn{1}{c}{QNLI}& \multicolumn{1}{c}{RTE}  & \multicolumn{1}{c}{SST2} & \multicolumn{1}{c}{IMDB} & \multicolumn{1}{c}{MRPC} & \multicolumn{1}{c}{QNLI}& \multicolumn{1}{c}{RTE}  & \multicolumn{1}{c}{SST2} & \multicolumn{1}{c}{IMDB} \\

\cmidrule(lr){1-11}

\multicolumn{1}{l}{\textit{CleanOnly}} & 56.6 & 64.9 & 33.9 & 40.4 & 51.3   & 60.5 & 59.8 & 39.5 & 40.2 & 55.6  \\ 

\cmidrule(lr){1-11}

\multicolumn{11}{l}{\textit{\textbf{Clean + DeepWordBug}}} \\

\multicolumn{1}{l}{\textit{AdvOnly}} & 73.3 & 73.6 & 62.8 & 68.4 & 56.3 & 75.4 & 70.4 & 61.3 & 66.8 & 56.9 \\ 

\multicolumn{1}{l}{\textit{AdvTrain}} & 66.8 & 70.5 & 48.1 & 43.3 & 53.1& 75.8 & 69.7  & 52.9 & 50.5 & 53.5 \\ 

\multicolumn{1}{l}{\textit{ModelSoup}}& {59.0} & {55.1} & {35.2} & {44.7} & {41.5}  & {52.9}  & {53.5} & \textbf{57.6} & {42.1} & {43.8} \\ 

\multicolumn{1}{l}{{\mymethod}} & \textbf{71.5} & \textbf{76.4} & \textbf{69.4} & \textbf{56.8} & \textbf{56.0} &  \textbf{78.6} & \textbf{69.3} & \textbf{72.8} &\textbf{56.9}& \textbf{54.9}\\ 

\cmidrule(lr){1-11}

\multicolumn{11}{l}{\textbf{Clean + TextBugger}} \\ 

\multicolumn{1}{l}{\textit{AdvOnly}} & 69.3 & 74.6 & 65.6 & 61.5 &51.4 & 72.9 & 73.0 & 60.9& 58.3 & 58.7  \\ 

\multicolumn{1}{l}{\textit{AdvTrain}} & 66.4 & 66.7 & 48.6 & 49.1 &48.9& 71.5 & 67.0 & 51.7 & 52.5 & 54.3 \\

\multicolumn{1}{l}{\textit{ModelSoup}} & {56.9} & {57.3} & {41.2} & {40.9} &{40.3}  & {63.2} & {54.6} & {41.8} & {43.0} & {44.5} \\ 

\multicolumn{1}{l}{{\mymethod}} & \textbf{68.2} & \textbf{71.4} & \textbf{59.8} & \textbf{59.7} & \textbf{52.5 }& \textbf{73.6} & \textbf{71.5} & \textbf{58.2} & \textbf{57.1} & \textbf{57.3} \\

\bottomrule
\end{tabular}
\caption{Cross Attack Evaluation (from Character-based to Word-based) with BERT (2)}
\label{cross_attack_character_word_bert_2}
\end{table*}

\paragraph{Between Word-based attacks.}
Tables \ref{cross_attack_wordbase_roberta_1}, \ref{cross_attack_wordbase_roberta_2}, \ref{cross_attack_wordbase_bert_1}, \ref{cross_attack_wordbase_bert_2} show detailed cross attack evaluations between Word-based methods on RoBERTa, BERT.

\renewcommand{\tabcolsep}{0.8pt}
\begin{table*}[htb!]
\footnotesize
\begin{tabular}{lcccccccccccccccccccc}
\toprule
\multicolumn{1}{c}{\multirow{2}{*}{\textit{\textbf{Attacker}}}} & \multicolumn{10}{c}{\textbf{PS}}  &  \multicolumn{10}{c}{\textbf{PW}}  \\

\cmidrule(lr){2-11} \cmidrule(lr){12-21}
\multicolumn{1}{c}{}   & \multicolumn{2}{c}{MRPC} & \multicolumn{2}{c}{QNLI}& \multicolumn{2}{c}{RTE}  & \multicolumn{2}{c}{SST2} & \multicolumn{2}{c}{IMDB} & \multicolumn{2}{c}{MRPC} & \multicolumn{2}{c}{QNLI}& \multicolumn{2}{c}{RTE}  & \multicolumn{2}{c}{SST2} & \multicolumn{2}{c}{IMDB} \\
\cmidrule(lr){2-3} \cmidrule(lr){4-5} \cmidrule(lr){6-7} \cmidrule(lr){8-9} \cmidrule(lr){10-11} \cmidrule(lr){12-13} \cmidrule(lr){14-15} \cmidrule(lr){16-17} \cmidrule(lr){18-19} \cmidrule(lr){20-21}

\multicolumn{1}{c}{}  & \multicolumn{1}{c}{Clean} & \multicolumn{1}{c}{Adv} & \multicolumn{1}{c}{Clean} & \multicolumn{1}{c}{Adv}& \multicolumn{1}{c}{Clean} & \multicolumn{1}{c}{Adv}& \multicolumn{1}{c}{Clean} & \multicolumn{1}{c}{Adv}& \multicolumn{1}{c}{Clean} & \multicolumn{1}{c}{Adv}& \multicolumn{1}{c}{Clean} & \multicolumn{1}{c}{Adv}& \multicolumn{1}{c}{Clean} & \multicolumn{1}{c}{Adv}& \multicolumn{1}{c}{Clean} & \multicolumn{1}{c}{Adv}& \multicolumn{1}{c}{Clean} & \multicolumn{1}{c}{Adv}& \multicolumn{1}{c}{Clean} & \multicolumn{1}{c}{Adv} \\
\cmidrule(lr){1-21}

\multicolumn{1}{l}{\textit{CleanOnly}} &  90.0&57.2 & 94.8&60.6 & 85.2&37.8 & 95.9&42.4 & 92.5&53.4 & &60.3 & &63.2 & &35.4 & &67.7 & &54.1 \\ 

\cmidrule(lr){1-21}

\multicolumn{21}{l}{\textit{\textbf{Clean + TextFooler}}} \\

\multicolumn{1}{l}{\textit{AdvOnly}} & 68.4&63.2 & 49.4&59.1 & 52.7&68.1 & 51.0&57.2 & 51.1&79.8   &  68.4&67.2 & 49.4&64.5 & 52.7&71.8 & 51.0&57.2 & 51.1&80.3 \\ 

\multicolumn{1}{l}{\textit{AdvTrain}}  & 87.8&60.7 & 92.0&56.4 & 84.5&60.7 & 95.2&52.0 & 90.6&71.1 & 87.8& 64.9 & 92.0&61.6 & 84.5&64.1 & 95.2&56.3 & 90.6&67.1\\
\bottomrule
\end{tabular}
\caption{Cross Attack Evaluation between Word-based methods on RoBERTa (1)}
\label{cross_attack_wordbase_roberta_1}
\end{table*}

\renewcommand{\tabcolsep}{0.8pt}
\begin{table*}[htb!]
\footnotesize
\begin{tabular}{lcccccccccccccccccccc}
\toprule
\multicolumn{1}{c}{\multirow{2}{*}{\textit{\textbf{Methods}}}} & \multicolumn{10}{c}{\textbf{TextFooler}} & \multicolumn{10}{c}{\textbf{BAE}}     \\

\cmidrule(lr){2-11} \cmidrule(lr){12-21}
\multicolumn{1}{c}{}   & \multicolumn{2}{c}{MRPC} & \multicolumn{2}{c}{QNLI}& \multicolumn{2}{c}{RTE}  & \multicolumn{2}{c}{SST2} & \multicolumn{2}{c}{IMDB} & \multicolumn{2}{c}{MRPC} & \multicolumn{2}{c}{QNLI}& \multicolumn{2}{c}{RTE}  & \multicolumn{2}{c}{SST2} & \multicolumn{2}{c}{IMDB} \\
\cmidrule(lr){2-3} \cmidrule(lr){4-5} \cmidrule(lr){6-7} \cmidrule(lr){8-9} \cmidrule(lr){10-11} \cmidrule(lr){12-13} \cmidrule(lr){14-15} \cmidrule(lr){16-17} \cmidrule(lr){18-19} \cmidrule(lr){20-21}

\multicolumn{1}{c}{}  & \multicolumn{1}{c}{Clean} & \multicolumn{1}{c}{Adv} & \multicolumn{1}{c}{Clean} & \multicolumn{1}{c}{Adv}& \multicolumn{1}{c}{Clean} & \multicolumn{1}{c}{Adv}& \multicolumn{1}{c}{Clean} & \multicolumn{1}{c}{Adv}& \multicolumn{1}{c}{Clean} & \multicolumn{1}{c}{Adv}& \multicolumn{1}{c}{Clean} & \multicolumn{1}{c}{Adv}& \multicolumn{1}{c}{Clean} & \multicolumn{1}{c}{Adv}& \multicolumn{1}{c}{Clean} & \multicolumn{1}{c}{Adv}& \multicolumn{1}{c}{Clean} & \multicolumn{1}{c}{Adv}& \multicolumn{1}{c}{Clean} & \multicolumn{1}{c}{Adv} \\
\cmidrule(lr){1-21}

\multicolumn{1}{l}{\textit{CleanOnly}} & 90.0&51.1 & 94.8&56.0 & 85.2&33.6 & 95.9&42.4 & 92.5&50.8  & 90.0&55.5 & 94.8&50.9 & 85.2&39.4 & 95.9&29.3 & 92.5&52.9 \\ 

\cmidrule(lr){1-21}

\multicolumn{21}{l}{\textit{\textbf{Clean + PW}}} \\

\multicolumn{1}{l}{\textit{AdvOnly}} & 68.2&58.6 & 52.9&90.5 & 51.3&68.9 & 50.9&55.5 & 54.6&60.6 & 68.2&65.8 & 52.9&86.6  & 51.3&68.6 & 50.9&44.8 & 54.6&79.4 \\ 

\multicolumn{1}{l}{\textit{AdvTrain}}  & 87.0&56.4 & 92.6&82.4 & 83.4&64.3 & 95.5&54.0 &91.6&59.2 &  87.0&61.6 & 92.6&78.9 & 83.4&63.1 & 95.5&41.4 & 91.6&68.6 \\
\bottomrule
\end{tabular}
\caption{Cross Attack Evaluation between Word-based methods on RoBERTa (2)}
\label{cross_attack_wordbase_roberta_2}
\end{table*}

\renewcommand{\tabcolsep}{0.8pt}
\begin{table*}[htb!]
\footnotesize
\begin{tabular}{lcccccccccccccccccccccc}
\toprule
\multicolumn{1}{c}{\multirow{2}{*}{\textit{\textbf{Methods}}}} & \multicolumn{10}{c}{\textbf{PS}}  &  \multicolumn{10}{c}{\textbf{PW}}  \\

\cmidrule(lr){2-11} \cmidrule(lr){12-21}
\multicolumn{1}{c}{}   & \multicolumn{2}{c}{MRPC} & \multicolumn{2}{c}{QNLI}& \multicolumn{2}{c}{RTE}  & \multicolumn{2}{c}{SST2} & \multicolumn{2}{c}{IMDB} & \multicolumn{2}{c}{MRPC} & \multicolumn{2}{c}{QNLI}& \multicolumn{2}{c}{RTE}  & \multicolumn{2}{c}{SST2} & \multicolumn{2}{c}{IMDB} \\
\cmidrule(lr){2-3} \cmidrule(lr){4-5} \cmidrule(lr){6-7} \cmidrule(lr){8-9} \cmidrule(lr){10-11} \cmidrule(lr){12-13} \cmidrule(lr){14-15} \cmidrule(lr){16-17} \cmidrule(lr){18-19} \cmidrule(lr){20-21}

\multicolumn{1}{c}{}  & \multicolumn{1}{c}{Clean} & \multicolumn{1}{c}{Adv} & \multicolumn{1}{c}{Clean} & \multicolumn{1}{c}{Adv}& \multicolumn{1}{c}{Clean} & \multicolumn{1}{c}{Adv}& \multicolumn{1}{c}{Clean} & \multicolumn{1}{c}{Adv}& \multicolumn{1}{c}{Clean} & \multicolumn{1}{c}{Adv}& \multicolumn{1}{c}{Clean} & \multicolumn{1}{c}{Adv}& \multicolumn{1}{c}{Clean} & \multicolumn{1}{c}{Adv}& \multicolumn{1}{c}{Clean} & \multicolumn{1}{c}{Adv}& \multicolumn{1}{c}{Clean} & \multicolumn{1}{c}{Adv}& \multicolumn{1}{c}{Clean} & \multicolumn{1}{c}{Adv} \\
\cmidrule(lr){1-21}

\multicolumn{1}{l}{\textit{CleanOnly}} & 83.3&56.6 & 90.5&64.9 & 65.0&33.9 & 92.5&40.4 & 88.9&51.3& 83.3&60.5 & 90.5&59.8 & 65.0&39.5 & 92.5&40.2 & 88.9&55.6  \\ 

\cmidrule(lr){1-21}

\multicolumn{21}{l}{\textit{\textbf{Clean + TextFooler}}} \\

\multicolumn{1}{l}{\textit{AdvOnly}} &  35.6&69.5 & 75.3&64.8& 46.9&60.7 & 23.6&63.3& 60.1&59.7 &  35.6&69.5 &75.3&63.9& 46.9&56.7 & 23.6&62.7 & 60.1&64.7 \\ 

\multicolumn{1}{l}{\textit{AdvTrain}}  & 78.7&66.8 & 86.8&60.3& 62.1&43.2 & 92.5&47.3 & 86.7&56.4 &  78.7&74.4 & 86.8&59.4 & 62.1&47.9 & 92.5&51.4 & 86.7&60.3 \\
\bottomrule
\end{tabular}
\caption{Cross Attack Evaluation between Word-based methods on BERT (1)}
\label{cross_attack_wordbase_bert_1}
\end{table*}

\renewcommand{\tabcolsep}{0.8pt}
\begin{table*}[htb!]
\footnotesize
\begin{tabular}{lccccccccccccccccccccc}
\toprule
\multicolumn{1}{c}{\multirow{2}{*}{\textit{\textbf{Methods}}}} & \multicolumn{10}{c}{\textbf{TextFooler}} & \multicolumn{10}{c}{\textbf{BAE}}     \\

\cmidrule(lr){2-11} \cmidrule(lr){12-21}
\multicolumn{1}{c}{}   & \multicolumn{2}{c}{MRPC} & \multicolumn{2}{c}{QNLI}& \multicolumn{2}{c}{RTE}  & \multicolumn{2}{c}{SST2} & \multicolumn{2}{c}{IMDB} & \multicolumn{2}{c}{MRPC} & \multicolumn{2}{c}{QNLI}& \multicolumn{2}{c}{RTE}  & \multicolumn{2}{c}{SST2} & \multicolumn{2}{c}{IMDB} \\
\cmidrule(lr){2-3} \cmidrule(lr){4-5} \cmidrule(lr){6-7} \cmidrule(lr){8-9} \cmidrule(lr){10-11} \cmidrule(lr){12-13} \cmidrule(lr){14-15} \cmidrule(lr){16-17} \cmidrule(lr){18-19} \cmidrule(lr){20-21}

\multicolumn{1}{c}{}  & \multicolumn{1}{c}{Clean} & \multicolumn{1}{c}{Adv} & \multicolumn{1}{c}{Clean} & \multicolumn{1}{c}{Adv}& \multicolumn{1}{c}{Clean} & \multicolumn{1}{c}{Adv}& \multicolumn{1}{c}{Clean} & \multicolumn{1}{c}{Adv}& \multicolumn{1}{c}{Clean} & \multicolumn{1}{c}{Adv}& \multicolumn{1}{c}{Clean} & \multicolumn{1}{c}{Adv}& \multicolumn{1}{c}{Clean} & \multicolumn{1}{c}{Adv}& \multicolumn{1}{c}{Clean} & \multicolumn{1}{c}{Adv}& \multicolumn{1}{c}{Clean} & \multicolumn{1}{c}{Adv}& \multicolumn{1}{c}{Clean} & \multicolumn{1}{c}{Adv} \\
\cmidrule(lr){1-21}

\multicolumn{1}{l}{\textit{Clean Only}} & 83.3&64.8 & 90.5&62.9 & 65.0&35.1 & 92.5&52.5 & 88.9&52.0 &  83.3&60.8 & 90.5&53.1 & 65.0&37.9 & 92.5&27.7 & 88.9&49.6 \\ 

\cmidrule(lr){1-21}

\multicolumn{21}{l}{\textit{\textbf{Clean + PW}}} \\

\multicolumn{1}{l}{\textit{AdvOnly}} & 54.7&73.1 & 76.6&92.8& 46.2&58.2 & 19.3&66.1 & 58.5&47.9&  54.7&54.8 & 76.6&89.4 & 46.2&59.4 & 19.3&74.8 & 58.5&50.8  \\ 

\multicolumn{1}{l}{\textit{AdvTrain}}  & 78.9&71.0 & 85.3&85.9 & 59.9&46.2 & 90.5&61.4 & 87.9&45.8 & 78.9&49.2 & 85.3&86.5 & 59.9&48.2 & 90.5&35.3 & 87.9&47.5 \\
\bottomrule
\end{tabular}
\caption{Cross Attack Evaluation between Word-based methods on BERT (2)}
\label{cross_attack_wordbase_bert_2}
\end{table*}

\paragraph{Between Character-based attacks.}
Tables \ref{cross_attack_character_base_roberta_1}, \ref{cross_attack_character_base_roberta_2}, 
\ref{cross_attack_character_base_bert_1} and \ref{cross_attack_character_base_bert_2} show detailed cross-attack evaluations between Word-based methods on RoBERTa, BERT.

\renewcommand{\tabcolsep}{0.8pt}
\begin{table*}[htb!]
\footnotesize
\begin{tabular}{lccccccccccc}
\toprule
\multicolumn{1}{c}{\multirow{2}{*}{\textit{\textbf{Methods}}}} & \multicolumn{10}{c}{\textbf{TextBugger}}     \\

\cmidrule(lr){2-11}
\multicolumn{1}{c}{}   & \multicolumn{2}{c}{MRPC} & \multicolumn{2}{c}{QNLI}& \multicolumn{2}{c}{RTE}  & \multicolumn{2}{c}{SST2} & \multicolumn{2}{c}{IMDB} \\

\cmidrule(lr){2-3} \cmidrule(lr){4-5} \cmidrule(lr){6-7} \cmidrule(lr){8-9} \cmidrule(lr){10-11}

\multicolumn{1}{c}{}  & \multicolumn{1}{c}{Clean} & \multicolumn{1}{c}{Adv} & \multicolumn{1}{c}{Clean} & \multicolumn{1}{c}{Adv}& \multicolumn{1}{c}{Clean} & \multicolumn{1}{c}{Adv}& \multicolumn{1}{c}{Clean} & \multicolumn{1}{c}{Adv}& \multicolumn{1}{c}{Clean} & \multicolumn{1}{c}{Adv}\\
\cmidrule(lr){1-11}

\multicolumn{1}{l}{\textit{CleanOnly}} & 90.0&65.8 & 94.8&65.5 & 85.2&47.3 & 95.9&60.3 & 92.5&61.0  \\ 

\cmidrule(lr){1-11}

\multicolumn{11}{l}{\textit{\textbf{Clean + DeepWordBug}}} \\

\multicolumn{1}{l}{\textit{AdvOnly}} & 66.4&71.3 & 55.4&72.7 & 51.6&65.6 & 39.9&65.0 & 50.0&80.2  \\ 

\multicolumn{1}{l}{\textit{AdvTrain}}  & 88.5&71.3 & 93.3&70.0 & 79.4&57.6 & 93.9&63.4 & 91.7&75.6 \\ 

\multicolumn{1}{l}{\textit{ModelSoup}} & {68.4}&{37.6} & {61.0}&{57.3} & 55.6&{57.9} & {84.6}&{56.8} & {78.0}&{65.1} \\ 

\multicolumn{1}{l}{{\mymethod}} & \textbf{90.6}&\textbf{72.8} & \textbf{94.7}&\textbf{71.6} & \textbf{84.9}&\textbf{65.5} & \textbf{96.5}&\textbf{66.9} & \textbf{92.4}&\textbf{79.5} \\ 

\bottomrule
\end{tabular}
\caption{Cross Attack Evaluation between Character-based methods on RoBERTa (1)}
\label{cross_attack_character_base_roberta_1}
\end{table*}

\renewcommand{\tabcolsep}{0.8pt}
\begin{table*}[htb!]
\footnotesize
\begin{tabular}{lcccccccccc}
\toprule
\multicolumn{1}{c}{\multirow{2}{*}{\textit{\textbf{Methods}}}} & \multicolumn{10}{c}{\textbf{DeepWordBug}}     \\

\cmidrule(lr){2-11}
\multicolumn{1}{c}{}   & \multicolumn{2}{c}{MRPC} & \multicolumn{2}{c}{QNLI}& \multicolumn{2}{c}{RTE}  & \multicolumn{2}{c}{SST2} & \multicolumn{2}{c}{IMDB} \\

\cmidrule(lr){2-3} \cmidrule(lr){4-5} \cmidrule(lr){6-7} \cmidrule(lr){8-9} \cmidrule(lr){10-11}

\multicolumn{1}{c}{}  & \multicolumn{1}{c}{Clean} & \multicolumn{1}{c}{Adv} & \multicolumn{1}{c}{Clean} & \multicolumn{1}{c}{Adv}& \multicolumn{1}{c}{Clean} & \multicolumn{1}{c}{Adv}& \multicolumn{1}{c}{Clean} & \multicolumn{1}{c}{Adv}& \multicolumn{1}{c}{Clean} & \multicolumn{1}{c}{Adv}\\
\cmidrule(lr){1-11}

\multicolumn{1}{l}{\textit{CleanOnly}} & 90.0&66.3 & 94.8&65.8 & 85.2&52.7 & 95.9&52.6 & 92.5&55.1 \\ 

\cmidrule(lr){1-11}

\multicolumn{11}{l}{\textit{\textbf{Clean + TextBugger}}} \\

\multicolumn{1}{l}{\textit{AdvOnly}} & 62.5&70.5 & 50.2&69.8 & 52.7&65.7 & 50.9&58.8 & 51.2&71.4 \\ 

\multicolumn{1}{l}{\textit{AdvTrain}}  & 88.2&67.6  & 94.5&67.6 & 80.9&61.1 & 90.7&56.7 & 90.9&66.4 \\ 

\multicolumn{1}{l}{\textit{ModelSoup}} & {77.2}&{48.4} & {56.4}&{66.8} & {61.0}&{47.3} & {75.2}&{45.6} & {74.6}&{59.3} \\ 

\multicolumn{1}{l}{{\mymethod}} & \textbf{89.7}&\textbf{69.9} & \textbf{95.2}&\textbf{71.8} & \textbf{85.0}&\textbf{67.5} & \textbf{96.3}&\textbf{60.3} & \textbf{92.4}&\textbf{70.6}\\ 

\bottomrule
\end{tabular}
\caption{Cross Attack Evaluation between Character-based methods on RoBERTa (2)}
\label{cross_attack_character_base_roberta_2}
\end{table*}

\renewcommand{\tabcolsep}{0.8pt}
\begin{table*}[htb!]
\footnotesize
\begin{tabular}{lcccccccccc}
\toprule
\multicolumn{1}{c}{\multirow{2}{*}{\textit{\textbf{Methods}}}} & \multicolumn{10}{c}{\textbf{TextBugger}}     \\

\cmidrule(lr){2-11}
\multicolumn{1}{c}{}   & \multicolumn{2}{c}{MRPC} & \multicolumn{2}{c}{QNLI}& \multicolumn{2}{c}{RTE}  & \multicolumn{2}{c}{SST2} & \multicolumn{2}{c}{IMDB} \\

\cmidrule(lr){2-3} \cmidrule(lr){4-5} \cmidrule(lr){6-7} \cmidrule(lr){8-9} \cmidrule(lr){10-11}

\multicolumn{1}{c}{}  & \multicolumn{1}{c}{Clean} & \multicolumn{1}{c}{Adv} & \multicolumn{1}{c}{Clean} & \multicolumn{1}{c}{Adv}& \multicolumn{1}{c}{Clean} & \multicolumn{1}{c}{Adv}& \multicolumn{1}{c}{Clean} & \multicolumn{1}{c}{Adv}& \multicolumn{1}{c}{Clean} & \multicolumn{1}{c}{Adv}\\
\cmidrule(lr){1-11}

\multicolumn{1}{l}{\textit{CleanOnly}} & 83.3&66.0 & 90.5&62.0 & 65.0&37.8 & 92.5&50.9 & 88.9&59.3 \\ 

\cmidrule(lr){1-11}

\multicolumn{11}{l}{\textit{\textbf{Clean + DeepWordBug}}} \\

\multicolumn{1}{l}{\textit{AdvOnly}} & 75.2&74.1 & 50.8&66.4 & 54.5&63.3 & 47.4&69.6 & 58.9&74.5 \\ 

\multicolumn{1}{l}{\textit{AdvTrain}}  & 81.6&77.7 & 90.6&62.7 & 61.4&56.2 & 91.0&66.6 & 87.2&71.4\\ 

\multicolumn{1}{l}{\textit{ModelSoup}} & {68.9}&{56.0} & {76.9}&{55.2} & {54.5}&{51.4} & {86.4}&{50.7} & {52.6}&{53.3} \\ 

\multicolumn{1}{l}{{\mymethod}} & \textbf{82.4}&\textbf{75.9} & \textbf{89.3}&\textbf{65.2} & \textbf{63.1}&\textbf{62.8} & \textbf{91.4}&\textbf{69.0} & \textbf{88.0}&\textbf{73.9} \\ 

\bottomrule
\end{tabular}
\caption{Cross Attack Evaluation between Character-based methods on BERT (1)}
\label{cross_attack_character_base_bert_1}
\end{table*}

\renewcommand{\tabcolsep}{0.8pt}
\begin{table*}[htb!]
\footnotesize
\begin{tabular}{lcccccccccc}
\toprule
\multicolumn{1}{c}{\multirow{2}{*}{\textit{\textbf{Methods}}}} & \multicolumn{10}{c}{\textbf{DeepWordBug}}     \\

\cmidrule(lr){2-11}
\multicolumn{1}{c}{}   & \multicolumn{2}{c}{MRPC} & \multicolumn{2}{c}{QNLI}& \multicolumn{2}{c}{RTE}  & \multicolumn{2}{c}{SST2} & \multicolumn{2}{c}{IMDB} \\

\cmidrule(lr){2-3} \cmidrule(lr){4-5} \cmidrule(lr){6-7} \cmidrule(lr){8-9} \cmidrule(lr){10-11}

\multicolumn{1}{c}{}  & \multicolumn{1}{c}{Clean} & \multicolumn{1}{c}{Adv} & \multicolumn{1}{c}{Clean} & \multicolumn{1}{c}{Adv}& \multicolumn{1}{c}{Clean} & \multicolumn{1}{c}{Adv}& \multicolumn{1}{c}{Clean} & \multicolumn{1}{c}{Adv}& \multicolumn{1}{c}{Clean} & \multicolumn{1}{c}{Adv}\\
\cmidrule(lr){1-11}

\multicolumn{1}{l}{\textit{CleanOnly}} &  83.3&51.3 & 90.5&52.5 & 65.0&33.6 & 92.5&40.3 & 88.9&52.2 \\ 

\cmidrule(lr){1-11}

\multicolumn{11}{l}{\textit{\textbf{Clean + TextBugger}}} \\

\multicolumn{1}{l}{\textit{AdvOnly}} & 38.7&80.9 & 52.1&67.6 & 45.1&68.1 & 23.2&69.7 & 53.9&75.3 \\ 

\multicolumn{1}{l}{\textit{AdvTrain}}  & 80.6&77.4 & 90.4&65.4 & 57.8&54.6 & 93.5&62.1 & 88.7&72.2 \\ 

\multicolumn{1}{l}{\textit{ModelSoup}} & {70.3}&{69.4} & {81.5}&{51.6} & {54.5}&{66.4} & {74.9}&{34.6} & {53.3}&{58.7} \\ 

\multicolumn{1}{l}{{\mymethod}} & \textbf{82.0}&\textbf{80.1} & \textbf{90.0}&\textbf{66.9} & \textbf{59.8}&\textbf{67.5} & \textbf{92.0}&\textbf{65.8} & \textbf{88.7}&\textbf{75.1}  \\ 

\bottomrule
\end{tabular}
\caption{Cross Attack Evaluation between Character-based methods on BERT (2)}
\label{cross_attack_character_base_bert_2}
\end{table*}

\subsection{Detailed Cross Attack Evaluation when know m>1 adversarial attacks}
\label{defense_against_more_than_one_adversarial_attack}

Tables from \ref{cross_attack_word_character_roberta_m_2_t1} to \ref{cross_attack_word_character_roberta_m_3_t2} show average model generalization and adversarial robustness across tasks when utilized in more than 1 adversarial attack.

\renewcommand{\tabcolsep}{0.8pt}
\begin{table*}[htb!]
\footnotesize
\begin{tabular}{llcccccccccccccccccccc}
\toprule
&\multicolumn{1}{c}{\multirow{2}{*}{\textit{\textbf{Methods}}}} & \multicolumn{10}{c}{\textbf{(TF, PW)$\rightarrow$BAE}}   & \multicolumn{10}{c}{\textbf{(TF, PW)$\rightarrow$PS}}    \\

\cmidrule(lr){3-12} \cmidrule(lr){13-22} 
& \multicolumn{1}{c}{}  & \multicolumn{2}{c}{MRPC} & \multicolumn{2}{c}{QNLI}& \multicolumn{2}{c}{RTE}  & \multicolumn{2}{c}{SST2} & \multicolumn{2}{c}{IMDB}  & \multicolumn{2}{c}{MRPC} & \multicolumn{2}{c}{QNLI}& \multicolumn{2}{c}{RTE}  & \multicolumn{2}{c}{SST2} & \multicolumn{2}{c}{IMDB} \\

\cmidrule(lr){3-4} \cmidrule(lr){5-6} \cmidrule(lr){7-8} \cmidrule(lr){9-10} \cmidrule(lr){11-12} \cmidrule(lr){13-14} \cmidrule(lr){15-16} \cmidrule(lr){17-18} \cmidrule(lr){19-20} \cmidrule(lr){21-22}

& \multicolumn{1}{c}{}  & \multicolumn{1}{c}{Clean} & \multicolumn{1}{c}{Adv} & \multicolumn{1}{c}{Clean} & \multicolumn{1}{c}{Adv}& \multicolumn{1}{c}{Clean} & \multicolumn{1}{c}{Adv}& \multicolumn{1}{c}{Clean} & \multicolumn{1}{c}{Adv}& \multicolumn{1}{c}{Clean} & \multicolumn{1}{c}{Adv}& \multicolumn{1}{c}{Clean} & \multicolumn{1}{c}{Adv}& \multicolumn{1}{c}{Clean} & \multicolumn{1}{c}{Adv}& \multicolumn{1}{c}{Clean} & \multicolumn{1}{c}{Adv}& \multicolumn{1}{c}{Clean} & \multicolumn{1}{c}{Adv}& \multicolumn{1}{c}{Clean} & \multicolumn{1}{c}{Adv} \\

\cmidrule(lr){1-22}

\multirow{6}{*}{\begin{sideways} \textit{\textbf{RoBERTa}} \end{sideways}} 

&\multicolumn{1}{l}{\textit{CleanOnly}} &  90.0 & 55.5 & 94.8 & 50.9  & 85.2 & 39.4 & 95.9 & 29.3 & 92.5 & 52.9 & 90.0 & 57.2 & 94.8 & 60.6 & 85.2 & 37.8 & 95.9 & 42.4 & 92.5 & 53.4  \\ 

&\multicolumn{1}{l}{\textit{AdvOnly}} & 68.4 & 65.8 & 49.5 & 85.2 & 52.7& 73.1& 49.1 & 48.3&52.4 & 79.9& 68.4 & 63.9 & 49.5 & 59.5 & 52.7 & 69.2 & 49.1 & 57.9 & 52.4 & 82.0\\

&\multicolumn{1}{l}{\textit{AdvTrain}} & 86.2 & 62.5 & 91.5 & 79.3 & 81.9 & 65.2 & 92.3 & 42.6 & 90.2 & 70.1 & 86.2 & 60.9 & 91.5 & 56.9 & 81.9 & 62.5  & 92.3 & 53.1 & 90.2 & 73.5 \\ 

&\multicolumn{1}{l}{\textit{ModelSoup}} & 51.6 & 49.3 & 81.5 & 63.7 & 60.4 & 52.9 & 76.8 & 43.2 & 80.5 & 54.2 & 51.6 & 55.5 & 81.5 & 54.2 & 60.4 & 56.3 & 76.8 & 43.6 & 80.5 & 46.9 \\ 

&\multicolumn{1}{l}{{\mymethod}} & 89.7 & 67.0 & 94.5 & 84.8 & 85.7 & 73.9 & 95.3 & 54.8 & 92.0 & 77.9 & 90.2 & 64.2 & 94.6 & 61.8 & 85.0 & 68.9 & 96.4 & 58.3 & 92.4 & 78.1 \\ 

\cmidrule(lr){1-22}

\multirow{6}{*}{\begin{sideways} \textit{\textbf{BERT}} \end{sideways}} 

&\multicolumn{1}{l}{\textit{CleanOnly}} &  83.3 & 60.8 & 90.5 & 53.1 & 65.0 & 37.9 & 92.5 & 27.7 & 88.9 & 49.6 & 83.3 & 56.6 & 90.5 & 64.9 & 65.0 & 33.9 & 92.5 & 40.4& 88.9 & 51.3     \\ 

&\multicolumn{1}{l}{\textit{AdvOnly}} & 41.7 & 67.1 &70.5 & 89.6& 44.8& 60.8& 18.7& 75.3 & 52.1 & 55.6& 41.7 & 70.0 & 70.5 & 65.2 &  44.8 & 62.7 & 18.7 & 70.2 & 52.1 & 64.3\\

&\multicolumn{1}{l}{\textit{AdvTrain}} & 78.0 & 62.8 & 83.9 & 86.5 & 58.6 & 50.3 & 88.5 & 40.9 & 86.3 & 52.2 & 78.0 & 69.0 & 83.9 & 61.9 & 58.6 & 44.5 & 88.5 & 49.5& 86.3 & 56.5 \\ 

&\multicolumn{1}{l}{\textit{ModelSoup}} & 64.3 & 53.4 & 77.3 & 60.2 & 43.7 & 33.8 & 63.5 & 34.9 & 63.4 & 43.8 & 64.3 & 46.9 & 77.3 & 53.8 & 43.7 & 39.9 & 63.5 & 44.3 & 63.4 & 44.5\\ 

&\multicolumn{1}{l}{{\mymethod}} & 82.0 & 54.3 & 88.1 & 90.2 & 63.7 & 56.1 & 91.2 & 72.9 & 88.5 & 49.3 & 82.7 & 69.0 & 89.5 & 66.9 & 64.5 & 57.3 & 92.5 & 64.3 & 88.0 & 61.8\\ 

\bottomrule
\end{tabular}
\caption{Cross Evaluation between Word-based attacks when knowing 2 adversarial attacks}
\label{cross_attack_word_character_roberta_m_2_t1}
\end{table*}

\renewcommand{\tabcolsep}{0.8pt}
\begin{table*}[htb!]
\footnotesize
\begin{tabular}{llcccccccccccccccccccc}
\toprule
&\multicolumn{1}{c}{\multirow{2}{*}{\textit{\textbf{Methods}}}} & \multicolumn{10}{c}{\textbf{(BAE, PS)$\rightarrow$TF}}   & \multicolumn{10}{c}{\textbf{(BAE, PS)$\rightarrow$PW}}    \\

\cmidrule(lr){3-12} \cmidrule(lr){13-22} 
& \multicolumn{1}{c}{}  & \multicolumn{2}{c}{MRPC} & \multicolumn{2}{c}{QNLI}& \multicolumn{2}{c}{RTE}  & \multicolumn{2}{c}{SST2} & \multicolumn{2}{c}{IMDB}  & \multicolumn{2}{c}{MRPC} & \multicolumn{2}{c}{QNLI}& \multicolumn{2}{c}{RTE}  & \multicolumn{2}{c}{SST2} & \multicolumn{2}{c}{IMDB} \\

\cmidrule(lr){3-4} \cmidrule(lr){5-6} \cmidrule(lr){7-8} \cmidrule(lr){9-10} \cmidrule(lr){11-12} \cmidrule(lr){13-14} \cmidrule(lr){15-16} \cmidrule(lr){17-18} \cmidrule(lr){19-20} \cmidrule(lr){21-22}

& \multicolumn{1}{c}{}  & \multicolumn{1}{c}{Clean} & \multicolumn{1}{c}{Adv} & \multicolumn{1}{c}{Clean} & \multicolumn{1}{c}{Adv}& \multicolumn{1}{c}{Clean} & \multicolumn{1}{c}{Adv}& \multicolumn{1}{c}{Clean} & \multicolumn{1}{c}{Adv}& \multicolumn{1}{c}{Clean} & \multicolumn{1}{c}{Adv}& \multicolumn{1}{c}{Clean} & \multicolumn{1}{c}{Adv}& \multicolumn{1}{c}{Clean} & \multicolumn{1}{c}{Adv}& \multicolumn{1}{c}{Clean} & \multicolumn{1}{c}{Adv}& \multicolumn{1}{c}{Clean} & \multicolumn{1}{c}{Adv}& \multicolumn{1}{c}{Clean} & \multicolumn{1}{c}{Adv} \\

\cmidrule(lr){1-22}

\multirow{6}{*}{\begin{sideways} \textit{\textbf{RoBERTa}} \end{sideways}} 

&\multicolumn{1}{l}{\textit{CleanOnly}} & \multicolumn{1}{c}{90.0}  & \multicolumn{1}{c}{51.1}  & \multicolumn{1}{c}{94.8}  & \multicolumn{1}{c}{56.0} & \multicolumn{1}{c}{85.2}  & \multicolumn{1}{c}{33.6} & \multicolumn{1}{c}{95.9}  & \multicolumn{1}{c}{42.4} & \multicolumn{1}{c}{92.5}  & 50.8 & \multicolumn{1}{c}{90.0}  & \multicolumn{1}{c}{60.3}   & \multicolumn{1}{c}{94.8}  & \multicolumn{1}{c}{63.2}  & \multicolumn{1}{c}{85.2}   & \multicolumn{1}{c}{35.4}   & \multicolumn{1}{c}{95.9}   & \multicolumn{1}{c}{67.7}  & \multicolumn{1}{c}{92.5}  & 54.1  \\ 

&\multicolumn{1}{l}{\textit{AdvOnly}} & 68.4 & 59.6 & 70.5& 68.2 & 52.1 & 76.8 & 49.1 & 54.5& 45.8 & 62.4& 68.4 & 67.4 & 70.5 & 70.3 & 52.1 & 71.8 & 49.1 & 59.0 & 45.8 & 80.3\\

&\multicolumn{1}{l}{\textit{AdvTrain}} & 86.2 & 56.6 & 91.0 & 67.5 & 80.6 & 59.8 & 92.5 & 52.3 & 89.3 & 59.8 & 86.2 & 65.1 & 91.0 & 64.9 & 80.6 & 68.3 & 92.5 & 50.5 & 89.3 & 66.8  \\ 

&\multicolumn{1}{l}{\textit{ModelSoup}} & 67.3 & 44.3 & 72.1 & 46.8 & 50.4 & 52.5 & 65.4 & 44.9 & 63.9 & 39.8  & 67.3 & 50.3 & 72.1 & 50.3 & 50.4 & 49.6 & 65.4 & 44.2 & 63.9 & 50.4\\ 

&\multicolumn{1}{l}{{\mymethod}} & 89.9 & 58.7 & 94.6 & 67.8 & 85.1 & 75.1 & 96.0 & 53.1 & 92.9 & 61.3 & 89.9 & 68.0 & 94.4 & 66.3 & 85.0 & 74.5 & 95.9 & 55.6 & 92.1 & 78.9  \\ 

\cmidrule(lr){1-22}

\multirow{6}{*}{\begin{sideways} \textit{\textbf{BERT}} \end{sideways}} 

&\multicolumn{1}{l}{\textit{CleanOnly}} & 83.3  & 64.8 & 90.5 & 62.9 & 65.0 & 35.1 & 92.5 & 52.5 & 88.9 & 52.0 & 83.3 &60.5& 90.5 & 59.8 & 65.0 & 39.5& 92.5 & 40.2 & 88.9 & 55.6 \\ 

&\multicolumn{1}{l}{\textit{AdvOnly}} & 44.1 & 66.5 & 40.7& 64.9& 43.7 & 59.3& 18.8 & 59.9& 51.2& 56.7& 44.1 & 68.3 & 40.7 & 47.7 & 43.7 & 64.6 & 18.8 & 59.4 & 51.2 & 66.8\\

&\multicolumn{1}{l}{\textit{AdvTrain}} & 79.4 & 72.8 &  88.4 & 55.8 & 61.8 & 52.3 & 91.5 & 57.1 & 84.2 & 53.4 & 79.4 & 61.5 & 88.4 & 52.4 & 61.8 & 53.3 & 91.5 & 49.8 & 84.2 & 62.7 \\ 

&\multicolumn{1}{l}{\textit{ModelSoup}} & 60.3 & 67.4 & 80.2 & 46.8 & 50.8 & 31.1 & 64.3 & 43.8 & 56.4 & 49.8 & 60.3 & 55.7 & 80.2 & 35.9 & 50.8 & 41.3 & 64.3 & 39.8 & 56.4 & 43.8  \\ 

&\multicolumn{1}{l}{{\mymethod}} & 82.1 & 71.3 & 90.0 & 58.5 & 65.5 & 54.9 & 92.6 & 57.8 & 88.2 & 55.8 & 83.1 & 64.0 & 90.2 & 46.4 & 63.1 & 58.2 & 92.4 & 53.2 & 88.0 & 64.9\\ 

\bottomrule
\end{tabular}
\caption{Cross Evaluation between Word-based attacks when knowing 2 adversarial attacks}
\label{cross_attack_word_character_roberta_m_2_t2}
\end{table*}

\renewcommand{\tabcolsep}{0.8pt}
\begin{table*}[htb!]
\footnotesize
\begin{tabular}{llcccccccccccccccccccc}
\toprule
&\multicolumn{1}{c}{\multirow{2}{*}{\textit{\textbf{Methods}}}} & \multicolumn{10}{c}{\textbf{(TF, PW, BAE)$\rightarrow$PS}}   & \multicolumn{10}{c}{\textbf{(PW, BAE, PS)$\rightarrow$TF}}    \\

\cmidrule(lr){3-12} \cmidrule(lr){13-22} 
& \multicolumn{1}{c}{}  & \multicolumn{2}{c}{MRPC} & \multicolumn{2}{c}{QNLI}& \multicolumn{2}{c}{RTE}  & \multicolumn{2}{c}{SST2} & \multicolumn{2}{c}{IMDB}  & \multicolumn{2}{c}{MRPC} & \multicolumn{2}{c}{QNLI}& \multicolumn{2}{c}{RTE}  & \multicolumn{2}{c}{SST2} & \multicolumn{2}{c}{IMDB} \\

\cmidrule(lr){3-4} \cmidrule(lr){5-6} \cmidrule(lr){7-8} \cmidrule(lr){9-10} \cmidrule(lr){11-12} \cmidrule(lr){13-14} \cmidrule(lr){15-16} \cmidrule(lr){17-18} \cmidrule(lr){19-20} \cmidrule(lr){21-22}

& \multicolumn{1}{c}{}  & \multicolumn{1}{c}{Clean} & \multicolumn{1}{c}{Adv} & \multicolumn{1}{c}{Clean} & \multicolumn{1}{c}{Adv}& \multicolumn{1}{c}{Clean} & \multicolumn{1}{c}{Adv}& \multicolumn{1}{c}{Clean} & \multicolumn{1}{c}{Adv}& \multicolumn{1}{c}{Clean} & \multicolumn{1}{c}{Adv}& \multicolumn{1}{c}{Clean} & \multicolumn{1}{c}{Adv}& \multicolumn{1}{c}{Clean} & \multicolumn{1}{c}{Adv}& \multicolumn{1}{c}{Clean} & \multicolumn{1}{c}{Adv}& \multicolumn{1}{c}{Clean} & \multicolumn{1}{c}{Adv}& \multicolumn{1}{c}{Clean} & \multicolumn{1}{c}{Adv} \\

\cmidrule(lr){1-22}

\multirow{6}{*}{\begin{sideways} \textit{\textbf{RoBERTa}} \end{sideways}} 

&\multicolumn{1}{l}{\textit{CleanOnly}} &  90.0 & 57.2  & 94.8 & 60.6  & 85.2 & 37.8  & 95.9 & 42.4  & 92.5 & 53.4 & \multicolumn{1}{c}{90.0}  & \multicolumn{1}{c}{51.1}  & \multicolumn{1}{c}{94.8}  & \multicolumn{1}{c}{56.0} & \multicolumn{1}{c}{85.2}  & \multicolumn{1}{c}{33.6} & \multicolumn{1}{c}{95.9}  & \multicolumn{1}{c}{42.4} & \multicolumn{1}{c}{92.5}  & 50.8   \\ 

&\multicolumn{1}{l}{\textit{AdvOnly}} & 68.4 & 64.2 & 49.1 & 60.1 & 52.0 & 70.4 & 50.3 & 59.3 & 48.2 &82.0&  68.5 & 59.5 & 71.6 & 70.7 & 52.7 & 74.3 & 50.8 & 56.5 & 45.4 & 63.0\\

&\multicolumn{1}{l}{\textit{AdvTrain}} & 86.2 & 62.5 & 90.2 & 57.8 & 81.9 & 64.8 & 91.2 & 55.7 & 86.3 & 78.4 & 85.1 & 57.3 & 88.5 & 69.1 & 78.4 & 63.4 & 90.1 & 54.0 & 86.8 & 61.7\\ 

&\multicolumn{1}{l}{\textit{ModelSoup}} & 78.3 & 53.2 & 59.4 & 51.6 & 61.4 & 57.4 & 54.3 & 45.9 & 41.5 & 44.8 & 62.5 & 40.5 & 64.9 & 41.8 & 45.7 & 47.9 & 53.2 & 40.8 & 50.6 & 36.9 \\ 

&\multicolumn{1}{l}{{\mymethod}} & 89.9 & 64.8 & 94.5 & 62.5 & 85.1 & 69.0 & 96.1 & 59.0 & 92.0 & 80.2 & 90.4 & 59.6 & 94.2 & 72.5 & 85.0 & 76.2 & 95.6 & 55.2 & 92.6 & 62.9\\ 

\cmidrule(lr){1-22}

\multirow{6}{*}{\begin{sideways} \textit{\textbf{BERT}} \end{sideways}} 

&\multicolumn{1}{l}{\textit{CleanOnly}} &   83.3 & 56.6 & 90.5 & 64.9 & 65.0 & 33.9 & 92.5 & 40.4& 88.9 & 51.3  & 83.3  & 64.8 & 90.1 & 62.9 & 64.5 & 35.1 & 92.5 & 52.5 & 88.9 & 52.0  \\ 

&\multicolumn{1}{l}{\textit{AdvOnly}} & 43.6 & 69.7 & 70.5 & 66.8 & 46.2 & 65.3 & 17.7 & 67.6 &50.9&68.9& 57.6 & 70.3 & 58.4 & 64.3 & 43.3 & 57.7 & 17.1 & 60.2 & 51.0 & 57.1\\

&\multicolumn{1}{l}{\textit{AdvTrain}} & 75.0 & 69.8 & 83.1 & 64.2 & 58.1 & 50.4 & 88.4 & 51.2 & 82.5 & 59.4 & 76.3 & 73.0 & 87.0 & 57.4 & 60.2 & 53.9 & 87.4 & 57.7& 82.4 & 53.0 \\ 

&\multicolumn{1}{l}{\textit{ModelSoup}} & 50.6 & 42.8 & 60.4 & 45.8 & 38.6 & 40.3 & 54.2 & 45.8 & 51.4 & 41.8 & 50.9 & 54.8 & 58.3 & 44.9 & 45.9 & 32.5 & 60.2 & 41.9 & 50.6 & 44.5 \\ 

&\multicolumn{1}{l}{{\mymethod}} & 83.0 & 69.0 & 89.0 & 68.0 & 64.8 & 60.8 & 92.0 & 66.2 & 88.5 & 64.9 & 83.0 & 72.1 & 90.1 & 61.8 & 65.2 & 59.8 & 92.0 & 58.2 & 88.5 & 56.3\\ 

\bottomrule
\end{tabular}
\caption{Cross Evaluation between Word-based attacks when knowing 3 adversarial attacks}
\label{cross_attack_word_character_roberta_m_3_t1}
\end{table*}

\renewcommand{\tabcolsep}{0.8pt}
\begin{table*}[htb!]
\footnotesize
\begin{tabular}{llcccccccccccccccccccc}
\toprule
&\multicolumn{1}{c}{\multirow{2}{*}{\textit{\textbf{Methods}}}} & \multicolumn{10}{c}{\textbf{(TF, BAE, PS)$\rightarrow$PW}}   & \multicolumn{10}{c}{\textbf{(TF, PW, PS)$\rightarrow$BAE}}    \\

\cmidrule(lr){3-12} \cmidrule(lr){13-22} 
& \multicolumn{1}{c}{}  & \multicolumn{2}{c}{MRPC} & \multicolumn{2}{c}{QNLI}& \multicolumn{2}{c}{RTE}  & \multicolumn{2}{c}{SST2} & \multicolumn{2}{c}{IMDB}  & \multicolumn{2}{c}{MRPC} & \multicolumn{2}{c}{QNLI}& \multicolumn{2}{c}{RTE}  & \multicolumn{2}{c}{SST2} & \multicolumn{2}{c}{IMDB} \\

\cmidrule(lr){3-4} \cmidrule(lr){5-6} \cmidrule(lr){7-8} \cmidrule(lr){9-10} \cmidrule(lr){11-12} \cmidrule(lr){13-14} \cmidrule(lr){15-16} \cmidrule(lr){17-18} \cmidrule(lr){19-20} \cmidrule(lr){21-22}

& \multicolumn{1}{c}{}  & \multicolumn{1}{c}{Clean} & \multicolumn{1}{c}{Adv} & \multicolumn{1}{c}{Clean} & \multicolumn{1}{c}{Adv}& \multicolumn{1}{c}{Clean} & \multicolumn{1}{c}{Adv}& \multicolumn{1}{c}{Clean} & \multicolumn{1}{c}{Adv}& \multicolumn{1}{c}{Clean} & \multicolumn{1}{c}{Adv}& \multicolumn{1}{c}{Clean} & \multicolumn{1}{c}{Adv}& \multicolumn{1}{c}{Clean} & \multicolumn{1}{c}{Adv}& \multicolumn{1}{c}{Clean} & \multicolumn{1}{c}{Adv}& \multicolumn{1}{c}{Clean} & \multicolumn{1}{c}{Adv}& \multicolumn{1}{c}{Clean} & \multicolumn{1}{c}{Adv} \\

\cmidrule(lr){1-22}

\multirow{6}{*}{\begin{sideways} \textit{\textbf{RoBERTa}} \end{sideways}} 

&\multicolumn{1}{l}{\textit{CleanOnly}} & \multicolumn{1}{c}{90.0}  & \multicolumn{1}{c}{60.3}   & \multicolumn{1}{c}{94.8}  & \multicolumn{1}{c}{63.2}  & \multicolumn{1}{c}{85.2}   & \multicolumn{1}{c}{35.4}   & \multicolumn{1}{c}{95.9}   & \multicolumn{1}{c}{67.7}  & \multicolumn{1}{c}{92.5}  & 54.1   & 90.0    & 55.5 &  94.8 & 50.9 &  85.2 & 39.4 &  95.9 & 29.3 &  92.5 & 52.9   \\ 

&\multicolumn{1}{l}{\textit{AdvOnly}} & 68.4 & 67.2 & 49.5 & 74.5 & 52.7 & 72.6 & 50.3 & 61.7 & 53.9& 81.2 & 68.4 & 65.8 & 49.5 & 85.3 & 52.7 & 73.9 & 48.3 & 49.1 & 51.6 & 80.3\\

&\multicolumn{1}{l}{\textit{AdvTrain}} & 86.0 & 66.3 & 90.1 & 66.3 & 75.1 & 69.8 & 90.4 & 53.4 & 86.3 & 69.0 & 83.1 & 64.0 & 89.4 & 81.6 & 78.9 & 68.1 & 89.1 & 44.8 & 88.9 & 72.5  \\ 

&\multicolumn{1}{l}{\textit{ModelSoup}} & 61.2 &  42.7 & 66.1 & 43.8 & 56.3 & 43.2 & 70.6 & 51.0 & 66.2 & 44.2 & 45.7 & 43.6 & 76.3 & 54.9 & 55.8 & 50.8 & 72.1 & 40.3 & 67.6 & 51.9 \\ 

&\multicolumn{1}{l}{{\mymethod}} & 90.1 & 68.4 & 94.6 & 68.9 & 84.9 & 74.0 & 95.5 & 58.9 & 92.4 & 80.5 & 89.9 & 66.8 & 94.0 & 86.0 & 85.8 & 73.0 & 95.8 & 56.0 & 93.0 & 79.6 \\ 

\cmidrule(lr){1-22}

\multirow{6}{*}{\begin{sideways} \textit{\textbf{BERT}} \end{sideways}} 

&\multicolumn{1}{l}{\textit{CleanOnly}} & 83.3 &60.5& 90.5 & 59.8 & 65.0 & 39.5& 92.5 & 40.2 & 88.9 & 55.6 & 83.5 & 60.8& 90.7 & 53.1 & 65.0 &37.9& 92.5 & 27.7 & 88.9  & 49.6   \\ 

&\multicolumn{1}{l}{\textit{AdvOnly}} & 39.7 & 70.0 & 40.2& 48.6& 48.4 & 66.7 & 19.2 & 61.8 &50.5 &67.8 & 50.2 & 58.8 &68.4  &90.2 &42.6 & 62.7 & 18.7 &75.4 & 51.4 & 57.4\\

&\multicolumn{1}{l}{\textit{AdvTrain}} & 74.8 & 63.6 & 86.3 & 54.9 & 58.9 & 55.8 & 88.4 & 51.4 & 85.1 & 63.0 & 76.1 & 64.2 & 80.3 & 87.0 & 57.1 & 55.2 & 85.1 & 46.8 & 84.1 & 54.0 \\ 

&\multicolumn{1}{l}{\textit{ModelSoup}} & 56.8 & 53.9 & 56.8 & 36.8 & 41.9 & 45.9 & 60.5 & 34.7 & 44.7 & 40.7 & 60.1 & 45.6 & 64.9 & 54.8 & 40.8 & 31.2 & 56.8 & 30.8 & 54.3 & 41.7 \\ 

&\multicolumn{1}{l}{{\mymethod}} & 83.5 & 64.2 & 90.1 & 49.8 & 64.2 & 63.5 & 91.8 & 56.4 & 89.2 & 65.0 & 83.2 & 56.9 & 88.9 & 90.0 & 64.2 & 59.8 & 91.9 & 74.1 & 88.6 & 52.5  \\ 

\bottomrule
\end{tabular}
\caption{Cross Evaluation between Word-based attacks when knowing 3 adversarial attacks}
\label{cross_attack_word_character_roberta_m_3_t2}
\end{table*}

\subsection{Detail analysis on space and time complexity}
\label{detail_time_and_space_analysis}
Detail time and space analysis of different methods can be seen in Table \ref{detail_cost_of_training}.

\end{document}